\documentclass[10pt,twocolumn,letterpaper]{article}

\usepackage{iccv}
\usepackage{times}
\usepackage{epsfig}
\usepackage{graphicx}
\usepackage{amsmath}
\usepackage{amssymb}
\usepackage{booktabs}
\usepackage{caption}
\usepackage{subcaption}
\usepackage{multirow}

\usepackage[pagebackref=true,breaklinks=true,letterpaper=true,colorlinks,bookmarks=false]{hyperref}

\iccvfinalcopy 

\def\iccvPaperID{3857} 

\ificcvfinal\fi
\begin{document}

\title{\vspace{-2ex}SRM : A Style-based Recalibration Module for Convolutional Neural Networks}

\author{HyunJae Lee\\
Lunit Inc.\\
{\tt\small hjlee@lunit.io}
\and
Hyo-Eun Kim\\
Lunit Inc.\\
{\tt\small hekim@lunit.io}
\and
Hyeonseob Nam\\
Lunit Inc.\\
{\tt\small hsnam@lunit.io}
}

\maketitle

\begin{abstract}
Following the advance of style transfer with Convolutional Neural Networks (CNNs), the role of styles in CNNs has drawn growing attention from a broader perspective.
In this paper, we aim to fully leverage the potential of styles to improve the performance of CNNs in general vision tasks.
We propose a Style-based Recalibration Module (SRM), a simple yet effective architectural unit, which adaptively recalibrates intermediate feature maps by exploiting their styles.
SRM first extracts the style information from each channel of the feature maps by style pooling, then estimates per-channel recalibration weight via channel-independent style integration. By incorporating the relative importance of individual styles into feature maps, SRM effectively enhances the representational ability of a CNN.
The proposed module is directly fed into existing CNN architectures with negligible overhead.
We conduct comprehensive experiments on general image recognition as well as tasks related to styles, which verify the benefit of SRM over recent approaches such as Squeeze-and-Excitation (SE). To explain the inherent difference between SRM and SE, we provide an in-depth comparison of their representational properties.

  
\end{abstract}

\section{Introduction}

\begin{figure*}
\begin{center}
    \includegraphics[width=0.85\textwidth]{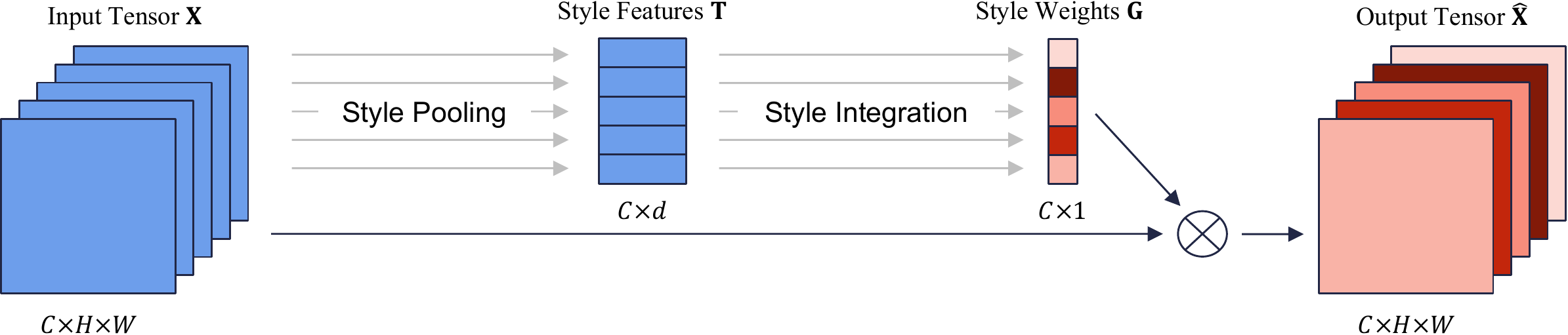}
    \caption{
    A Style-based Recalibration Module (SRM).
    SRM adaptively recalibrates input feature maps based on the style of an image via channel-independent style pooling and integration operators.}
    \label{fig:archi}
\end{center}
\end{figure*}

The evolution of convolutional neural networks (CNNs) has constantly pushed the boundaries of complex vision tasks~\cite{krizhevsky2012imagenet,liu2016ssd,chen2017deeplab}.
Besides their superior performance, a wide investigation has revealed that CNNs are capable of handling not only the content (i.e. shape) but also the style (i.e. texture) of an image.
Gatys et al.~\cite{gatys2015texture} discovered that the feature statistics of a CNN effectively encode the style information of an image, which laid the foundation of neural style transfer~\cite{gatys2016image,johnson2016perceptual,huang2017arbitrary}.
Recent approaches also pointed out that the styles play an unexpectedly significant role in the decision making process by standard CNNs~\cite{brendel2019approximating,geirhos2019imagenet}.
Furthermore, Karras et al.~\cite{karras2018style} demonstrated that a generative CNN architecture solely based on style manipulation achieves dramatic improvement in terms of realistic image generation.

Inspired by the tight link between the style and CNN representation, we aim to enhance the utilization of styles in a CNN to boost its representational power.
We propose a novel architectural unit, \textit{Style-based Recalibration Module} (SRM), which explicitly incorporates the styles into CNN representations through a form of feature recalibration.
Note that a CNN involves styles with varying levels of significance. While certain styles play an essential role, some are rather a nuisance factor to the task~\cite{nam2018batch}.
SRM dynamically estimates the relative importance of individual styles then reweights the feature maps based on the style importance, which allows the network to focus on meaningful styles while ignoring unnecessary ones.

The overall structure of SRM is illustrated in Figure~\ref{fig:archi}.
It consists of two main components: \textit{style pooling} and \textit{style integration}.
The style pooling operator extracts style features from each channel by summarizing feature responses across spatial dimensions. 
It is followed by the style integration operator, which produces example-specific style weights by utilizing the style features via channel-wise operation. 
The style weights finally recalibrate the feature maps to either emphasize or suppress their information.
Our proposed module is seamlessly integrated into modern CNN architecture and trained in an end-to-end manner. While SRM only imposes negligible additional parameters and computations, it remarkably improves the performance of the network.
Beyond the practical improvements, SRM provides an intuitive interpretation about the effect of channel-wise recalibration: it controls the contribution of styles by adjusting the global statistics of feature responses while maintaining their spatial configuration. 

Our experiments on image recognition \cite{russakovsky2015imagenet,krizhevsky2009learning} verify the effectiveness of SRM in general vision tasks. Throughout the experiment, SRM outperforms recent approaches~\cite{hu2018squeeze, hu2018gather} though it requires orders of magnitude less additional parameters.
Furthermore, we demonstrate the capability of SRM in arranging the contribution of styles. To this end, we conduct extensive experiments on style-related tasks such as classification with a texture-shape cue conflict~\cite{geirhos2019imagenet}, multi-domain classification~\cite{venkateswara2017deep}, texture recognition~\cite{cimpoi2016deep}, and style transfer~\cite{johnson2016perceptual}, where SRM brings exceptional performance improvements. 
We also provide comprehensive analysis and ablation studies to further investigate the behavior of SRM. 

The main contributions of this paper are as follows:
\begin{itemize}
  \item We present a style-based feature recalibration module which enhances the representational capability of a CNN by incorporating the styles into the feature maps.
  \item Despite its minimal overhead, the proposed module noticeably improves the performance of a network in general vision tasks as well as style-related tasks.
  \item Through in-depth analysis along with ablation study, we examine the internal behavior and validity of our method.
\end{itemize}

\section{Related Work}

\paragraph{Style Manipulation.} 
Manipulating the style information of CNNs has been widely studied in generative frameworks.
The pioneering work by Gatys et al.~\cite{gatys2016image} presented impressive style transfer results by exploiting the second-order statistics (i.e. the Gram matrix) of convolutional features as style representations.
Li et al.~\cite{li2017demystifying} also addressed style transfer by matching a variety of CNN feature statistics such as linear, polynomial and Gaussian kernels.
Adaptive instance normalization (AdaIN)~\cite{huang2017arbitrary} further showed that transferring channel-wise mean and standard deviation can efficiently change image styles. 
Recent work by Karras et al.~\cite{karras2018style} combined AdaIN into generative adversarial networks (GANs) to improve the generator by adjusting styles in intermediate layers.

The potential of styles in a CNN has been also investigated in discriminative settings.
BagNets~\cite{brendel2019approximating} demonstrated that a CNN constrained to rely on style information without considering spatial context performs surprisingly well on image classification.
Geirhos et al.~\cite{geirhos2019imagenet} discovered that CNNs (e.g. ImageNet-trained ResNet) are highly biased towards styles in their decision making process.
Batch-instance normalization~\cite{nam2018batch} achieved practical performance improvement by controlling styles, which learns static weights for individual styles and selectively normalizes unimportant ones.
In this work, we further facilitate the utilization of styles in designing a CNN architecture.
Our approach dynamically enriches feature representations by either highlighting or suppressing style regarding its relevance to the task.

\paragraph{Attention and Feature Recalibration.}
It is known that human pays attention to important parts of the visual input to better grasp the core information, rather than processing the whole visual signal at once~\cite{itti1998model,rensink2000dynamic,corbetta2002control}.
This mechanism has been extended to CNNs in a way of refining feature activations and showed effectiveness across a wide range of applications including object classification~\cite{jaderberg2015spatial,wang2017residual}, multimodal tasks~\cite{xu2015show,nam2017dual}, video classification~\cite{wang2018non}, etc.

More related to our work, Squeeze-and-Excitation (SE) ~\cite{hu2018squeeze} proposed a channel-wise recalibration operator that incorporates the interaction between channels.
It first aggregates the spatial information with global average pooling and captures the channel dependencies using a fully connected subnetwork.
Gather-Excite (GE)~\cite{hu2018gather} further explored this pipeline for better exploiting the global context with a convolutional aggregator.
Convolutional block attention module (CBAM)~\cite{woo2018cbam} also showed that the SE block can be improved by additionally utilizing max-pooled features and combining with a spatial attention module.
In contrast to the prior efforts, we reformulate channel-wise recalibration in terms of leveraging style information, without the aid of channel relationship nor spatial attention.
We present a style pooling approach which is superior to the standard global average or max pooling in our setting, as well as a channel-independent style integration method which is substantially more lightweight than fully connected counterparts yet more effective in various scenarios.
\section{Style-based Recalibration Module}

\begin{figure}[t]
\begin{center}
\setlength{\tabcolsep}{1em}
\begin{tabular}{cc}
\includegraphics[height=14em]{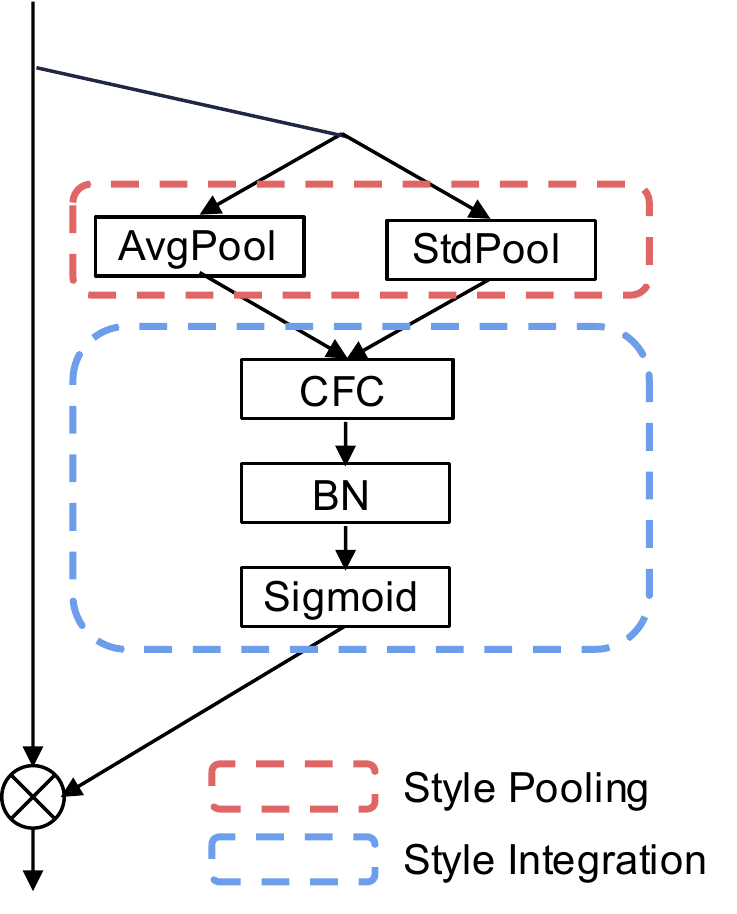}& \includegraphics[height=14em]{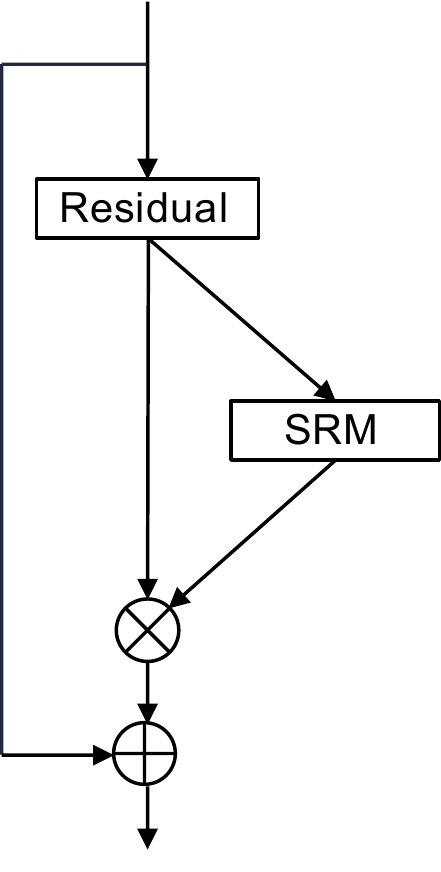}\\
(a) SRM & (b) Residual SRM
\end{tabular}
\caption{The schema of (a) SRM and (b) SRM integrated with a residual block. \textit{AvgPool} : global average pooling, \textit{StdPool} : global standard deviation pooling, \textit{CFC} : channel-wise fully connected layer, \textit{BN} : batch normalization.}
\label{fig:sam_se}
\end{center}
\end{figure}

Given an input tensor  $\mathbf{X} \in \mathbb{R}^{N\times C \times H \times W}$, SRM generates channel-wise recalibration weights $\mathbf{G} \in \mathbb{R}^{N \times C}$ based on the styles of $\mathbf{X}$, where $N$ indicates the number of examples in the mini-batch, $C$ is the number of channels; $H$ and $W$ indicate spatial dimensions.
It is divided into two sequential submodules: \textit{style pooling} for extracting an intermediate style representation $\mathbf{T} \in \mathbb{R}^{N \times C \times d}$ from $\mathbf{X}$, where $d$ is the number of style features, and \textit{style integtration} for estimating the style weights $\mathbf{G}$ from $\mathbf{T}$.
The final output $\hat{\mathbf{X}}$ is then computed by channel-wise multiplication between $\mathbf{G}$ and $\mathbf{X}$.
SRM is easily integrated into modern CNN architectures such as ResNets~\cite{he2016deep} and trained end-to-end.
Figure~\ref{fig:sam_se} illustrates the detailed structure of SRM and our configuration of the SRM integrated into a residual block.
 

\subsection{Style Pooling}
Extracting style information from intermediate convolutional feature maps has been widely studied in style transfer literature.
Motivated by \cite{huang2017arbitrary}, we adopt the channel-wise statistics---average and standard deviation---of each feature map as style features (i.e. $d=2$).  
Specifically, given input feature maps $\mathbf{X} \in \mathbb{R}^{N\times C \times H\times W}$, the style features $\mathbf{T} \in \mathbb{R}^{N\times C \times 2}$ are calculated by:
\begin{align}
\mu_{nc} &= \frac 1 {HW} \sum_{h=1}^H \sum_{w=1}^W x_{nchw}, \\
\sigma_{nc} &= \sqrt{ \frac 1 {HW} \sum_{h=1}^H \sum_{w=1}^W (x_{nchw} - \mu_{nc})^2}, \\
\mathbf{t}_{nc} &= [\mu_{nc}, \sigma_{nc}].
\end{align}
The style vector $\mathbf{t}_{nc} \in \mathbb{R}^2$ serves as a summary description of the style information for each example $n$ and channel $c$.
Other types of style features such as the correlations between different channels~\cite{gatys2016image} can be also included in the style vector, but we focus on the channel-wise statistics for efficiency and conceptual clarity. 
In section \ref{analysis}, we verify the practical benefits of the proposed style pooling compared to other approaches for gathering global information, e.g. using average pooling as in SE~\cite{hu2018squeeze} and additionally utilizing max pooling as in CBAM~\cite{woo2018cbam}.

\subsection{Style Integration}
The style features are converted into channel-wise style weights by a style integration operator.
The style weights are supposed to model the importance of the styles associated with individual channels so as to emphasize or suppress them accordingly.
To achieve this, we adopt a simple combination of a channel-wise fully connected (CFC) layer, a batch normalization (BN) layer, and a sigmoid activation function.
Given the style representation $\mathbf{T} \in \mathbb{R}^{N\times C \times 2}$ as an input, the style integration operator performs channel-wise encoding using learnable parameters $\mathbf{W} \in \mathbb{R}^{C \times 2}$:
\begin{align}
    z_{nc} = \mathbf{w}_{c} \cdot \mathbf{t}_{nc}
\end{align}
where $\mathbf{Z} \in \mathbb{R}^{N\times C}$ represents the encoded style features.
This operation can be viewed as a channel-independent fully connected layer with two input nodes and a single output, where the bias term is absorbed into the subsequent BN layer.
We then apply BN to facilitate training and a sigmoid function as a gating mechanism:
%
\begin{align}
\mu_{c}^{(z)} &= \frac 1 {N} \sum_{n=1}^N z_{nc}, \\
\sigma_{c}^{(z)} &= \sqrt{ \frac 1 {N} \sum_{n=1}^N (z_{nc} - \mu_{c}^{(z)})^2}, \\
\hat{z}_{nc} &= \gamma_{c} ( \frac {z_{nc} - \mu_{c}^{(z)}} {\sigma_{c}^{(z)}}) + \beta_{c}, \\
g_{nc} &= \frac 1 {1 + e^{-\hat{z}_{nc}}},
\end{align}
where $\mathbf{\gamma}, \mathbf{\beta} \in \mathbb{R}^{C}$ are affine transformation parameters, and $\mathbf{G} \in \mathbb{R}^{N\times C}$ represents the channel-wise style weights.
Note that BN makes use of fixed approximations of mean and variance at inference time, which allows the BN layer to be merged into the preceding CFC layer.
Consequently, the style integration for each channel boils down to a single CFC layer $f_{CFC}: \mathbb{R}^2 \rightarrow \mathbb{R}$ followed by an activation function $f_{ACT}: \mathbb{R} \rightarrow [0,1]$.
Finally, the original input $\mathbf{X}$ is recalibrated by the weights $\mathbf{G}$, so the output $\hat{\mathbf{X}} \in \mathbb{R}^{N\times C \times H\times W}$ is obtained by: 
%
\begin{align}
    \hat{\mathbf{x}}_{nc} &= g_{nc} \cdot \mathbf{x}_{nc}.
\end{align}

\begin{figure*}
\begin{center}
    \includegraphics[width=0.9\textwidth]{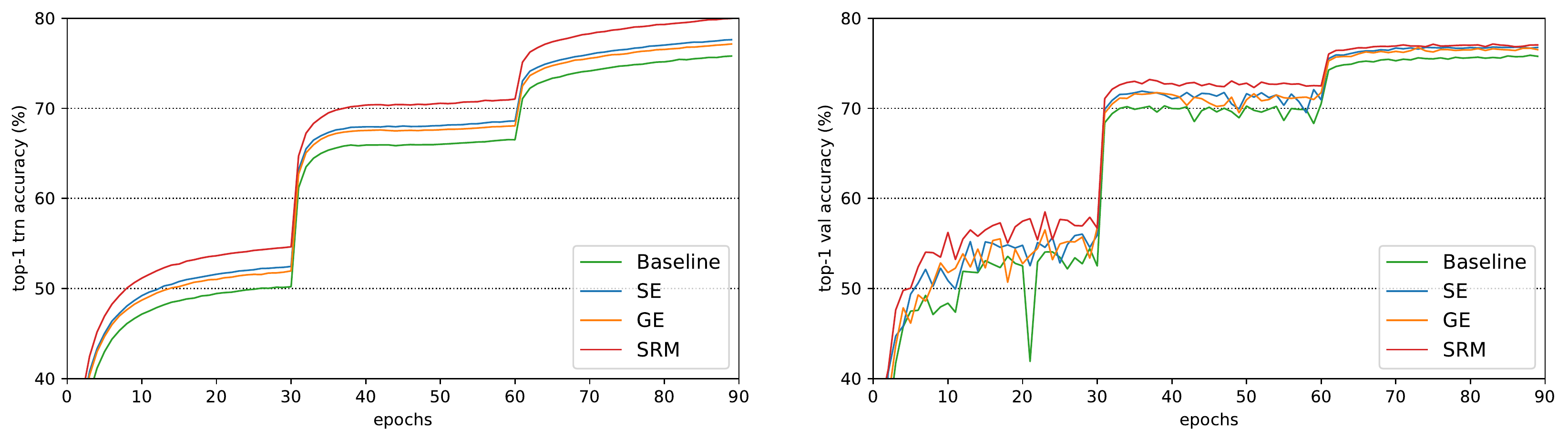}
    \caption{
    Training (left) and validation (right) curves on ImageNet-1K with ResNet-50 (baseline) and varying recalibration methods.}
    \label{fig:imagenet}
\end{center}
\end{figure*}

\subsection{Parameter and Computational Complexity}
SRM is designed to be lightweight in both terms of memory and computational complexity.
We first consider the additional parameters of SRM which come from the CFC and BN layers.
The number of parameters for each term is $\sum_{s=1}^S N_{s} \cdot C_{s} \cdot 2$ and $\sum_{s=1}^S N_{s} \cdot C_{s} \cdot 4$, respectively, where $S$ denotes the number of stages, $N_{s}$ is the the number of repeated blocks in $s$-th stage, and $C_{s}$ is the dimension of the output channels for $s$-th stage.
We follow the definition of \textit{stage} in \cite{hu2018squeeze} which refers to a group of convolutions with an identical spatial dimension.
In total, the number of extra parameters for SRM is:
\begin{align}
    6 \sum_{s=1}^S N_{s} \cdot C_{s}, 
\end{align}
which is typically negligible compared to SE's $\frac{2}{r} \sum_{s=1}^S N_{s} \cdot C_{s}^2$ where $r$ is its reduction ratio.
For instance, given ResNet-50 as a baseline architecture, SRM-ResNet-50 requires only 0.06M additional parameters whereas SE-ResNet-50 requires 2.53M.

In terms of computational complexity, SRM also introduces negligible extra computations to the original architecture.
For example, a single forward pass of a 224 $\times$ 224 pixel image for SRM-ResNet-50 requires additional 0.02 GFLOPs to ResNet-50 which requires 3.86 GFLOPs.
By adding only 0.52\% relative computational burden, SRM increases the top-1 validation accuracy of ResNet-50 from 75.89\% to 77.13\%,
which indicates that SRM offers a good trade-off between accuracy and efficiency.
\section{Experiment} \label{experiment}

In this section, we conduct a comprehensive evaluation across a wide range of problems and datasets to verify the effectiveness of SRM.
We re-implemented all competitors to compare under consistent settings for fair comparison. 

\subsection{Object Classification} \label{image_classification}

We first evaluate SRM on general object classification with ImageNet-1K~\cite{russakovsky2015imagenet} and CIFAR-10/100~\cite{krizhevsky2009learning}, in comparison with state-of-the-art methods such as Squeeze-and-Excitation (SE)~\cite{hu2018squeeze} and Gather-Excite (GE)\footnote{Among the several variants of GE, we compared with GE-$\theta$ which is mainly explored in their paper.}~\cite{hu2018gather}.
On the extension of \cite{brendel2019approximating,geirhos2019imagenet}, which suggest the crucial role of styles in the decision making by standard CNNs, we further demonstrate the potential of styles for improving the general performance of CNNs.

\paragraph{ImageNet-1K.}
The ImageNet-1K dataset~\cite{russakovsky2015imagenet} consists of 1,000 classes with 1.3 million training and 50,000 validation images.
We follow the standard practice for data augmentation and optimization \cite{he2016deep}.
The input images are randomly cropped to  224$\times$224 patches and random horizontal flipping is applied.
The networks are trained by SGD with a batch size of 256 on 8 GPUs, a momentum of 0.9, and a weight decay of 0.0001.
We train the networks for 90 epochs from the scratch with an initial learning rate of 0.1 which is divided by 10 every 30 epochs.
Single center crop evaluation is performed on 224$\times$224 patches where each image is first resized so that the shorter side is 256. 

Figure \ref{fig:imagenet} illustrates the training and validation curves of ResNet-50 with SRM and other feature recalibration methods.
Throughout the whole training process, SRM exhibits considerably higher accuracy than SE and GE on both training and validation curves.
This implies that utilizing styles with SRM is more effective than modeling channel interdependencies with SE or gathering global context with GE, in both terms of facilitating training and improving generalization.
Table~\ref{table:imagenet} also demonstrates that 
SRM significantly boosts the performance of the baseline architecture (ResNet-50/101) with almost the same number of parameters and computations.
On the other hand, due to its tendency of slow convergence as mentioned in~\cite{hu2018gather}, GE does not exhibit improved performance in a deeper network under a fixed-length training schedule. 
It is worth noting that SRM outperforms SE and GE with orders of magnitude less additional parameters.
For example, SE-ResNet-50 and GE-ResNet-50 require 2.53M and 5.56M additional parameters to ResNet-50, respectively, but SRM-ResNet-50 only requires 0.06M (2.37\% of SE and 1.08\% of GE) which shows the exceptional parameter efficiency of SRM.

\begin{table}
\caption{Top-1 and top-5 accuracy (\%) on the ImageNet-1K validation set and complexity comparison.}
\vspace{-1em}
\begin{center}
\begin{tabular}{c|c|c|c|c}
\hline
Model & Params & GFLOPs & top-1 & top-5 \\
\hline
ResNet-50 & 25.56M & 3.86 & 75.89 & 92.85 \\
SE-ResNet-50 & 28.09M & 3.87 & 76.80 & 93.39 \\
GE-ResNet-50 & 31.12M & 3.87 & 76.75 & 93.41 \\
SRM-ResNet-50 & 25.62M & 3.88 & \textbf{77.13}& \textbf{93.51} \\
\hline
ResNet-101 & 44.55M & 7.58 & 77.40 & 93.59 \\
SE-ResNet-101 & 49.33M & 7.60 & 78.08 & 93.95 \\
GE-ResNet-101 & 53.58M & 7.60 & 77.36 & 93.64 \\
SRM-ResNet-101 & 44.68M & 7.62 & \textbf{78.47} & \textbf{94.20}\\
\hline
\end{tabular}
\end{center}
\label{table:imagenet}
\end{table}

\begin{table}
\caption{Accuracy (\%) on the CIFAR-10/100 test sets with a ResNet-56 baseline and complexity comparison.}
\vspace{-1em}
\begin{center}
\begin{tabular}{c|c|c|c|c}
\hline
 & \multicolumn{2}{|c|}{CIFAR-10} & \multicolumn{2}{|c}{CIFAR-100} \\ \cline{2-5}
Model & Params & top-1 & Params & top-1 \\
\hline
Baseline & 0.87M &  93.77 & 0.89M & 74.76 \\
SE & 0.97M & 94.60 & 0.99M & 76.10  \\
GE & 1.91M & 94.32 & 1.94M & 76.02  \\
SRM & 0.89M  & \textbf{95.05} & 0.91M  & \textbf{76.93} \\
\hline
\end{tabular}
\end{center}
\label{table:cifar}
\end{table}

\paragraph{CIFAR-10/100.}
We also evaluate the performance of SRM on the CIFAR-10/100 dataset~\cite{krizhevsky2009learning} which consists of 50,000 training and 10,000 test images of 32$\times$32 pixels.
On the training phase, each image is zero-padded with 4 pixels then randomly cropped to the original size, and evaluation is performed on the original images.
The networks are trained with SGD for 64,000 iterations with a mini-batch size of 128 on a single GPU, a momentum of 0.9, and a weight decay of 0.0001. The initial learning rate is set to 0.2 which is divided by 10 at 32,000 and 48,000 iterations.
As presented in Table~\ref{table:cifar}, SRM considerably improves the accuracy on both CIFAR-10 and 100 with minimal parameter increases, which suggests that the effectiveness of SRM is not constrained to ImageNet.

\subsection{Style-Related Classification}

The proposed idea views channel-wise recalibration as an adjustment of intermediate styles, which is achieved by exploiting the global statistics of respective feature maps.
This interpretation motivates us to explore the effect of SRM on style-related tasks where explicitly manipulating style information could bring prominent benefits.

\begin{table}
\caption{Top-1 and top-5 accuracy (\%) on the validation sets of Stylized-ImageNet and ImageNet with a ResNet-50 baseline, when trained on Stylized-ImageNet.}
\vspace{-1em}
\begin{center}
\begin{tabular}{c|c|c|c|c}
\hline
 & \multicolumn{2}{|c|}{Stylized-ImageNet} & \multicolumn{2}{|c}{ImageNet} \\ \cline{2-5}
 & top-1 & top-5 & top-1 & top-5 \\
\hline
Bseline &\ \  53.93 \  & 76.75 & 56.11 & 79.17\\
SE & 58.31 & 80.80 & 60.15 & 82.54 \\
SRM & \textbf{60.69} & \textbf{82.56} & \textbf{62.12} & \textbf{84.06} \\
\hline
\end{tabular}
\end{center}
\label{table:stylized-imagenet}
\end{table}

\begin{table}
\caption{Accuracy (\%) on the Office-Home dataset with a ResNet-18 baseline, averaged over 5-fold cross validation.}
\vspace{-1em}
\begin{center}
\begin{tabular}{c|c|c|c|c|c}
\hline
  & Ar & Cl & Pr & Rw & Avg. \\
\hline
Baseline & 37.49 & 60.73 & 72.81 & 52.12 & 55.47 \\ 
SE & 39.55 & 62.75 & 75.60 & 55.52 & 58.36  \\ 
SRM & \textbf{40.50} & \textbf{64.97} & \textbf{76.12} & \textbf{56.30} & \textbf{59.47}  \\ 
\hline
\end{tabular}
\end{center}
\label{table:officehome}
\end{table}

\begin{figure*}
\newcommand{\addstyle}[1]{\includegraphics[width=6em, height=6em]{figure/style_transfer/#1.jpg}}
\begin{center}
\setlength{\tabcolsep}{0.05em}
\begin{tabular}{cccccc}
Style & Content & BN  & BN+SE & BN+SRM & IN \\
\addstyle{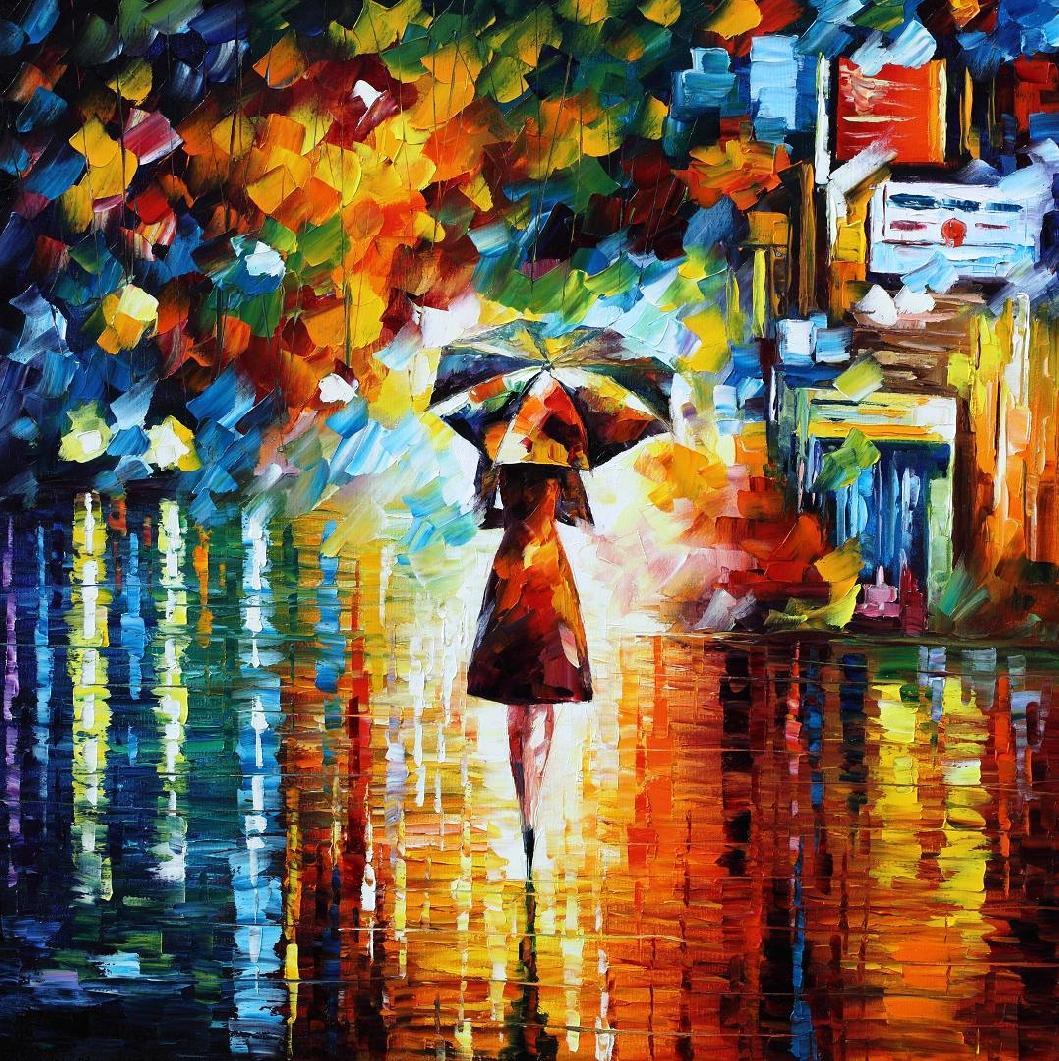} & \addstyle{rain-princess/content} & \addstyle{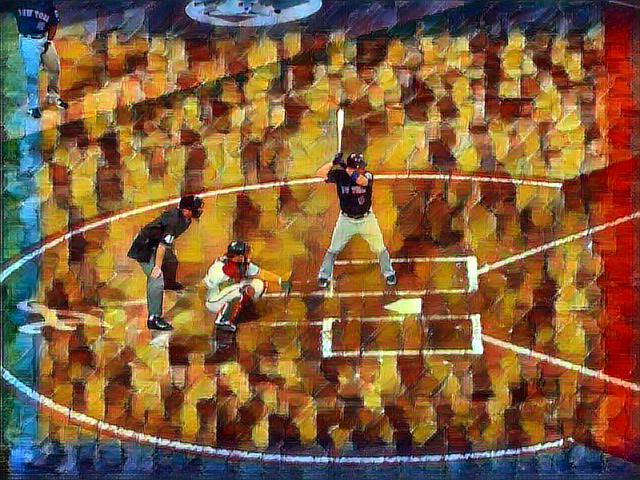}  & \addstyle{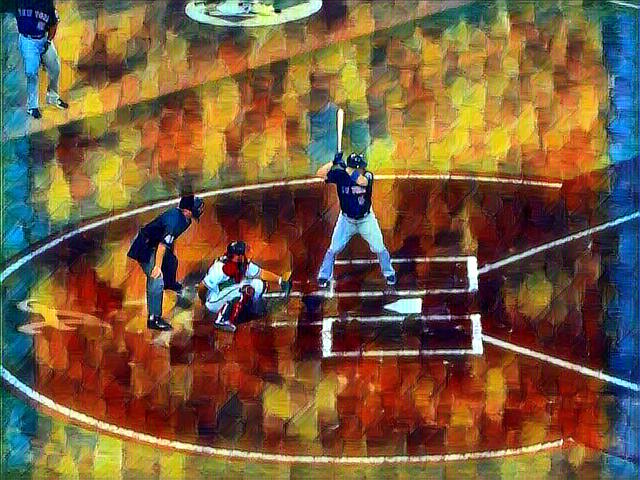} & \addstyle{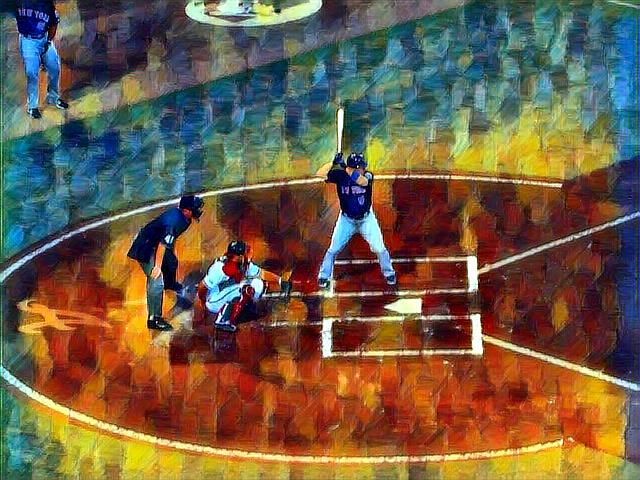} & \addstyle{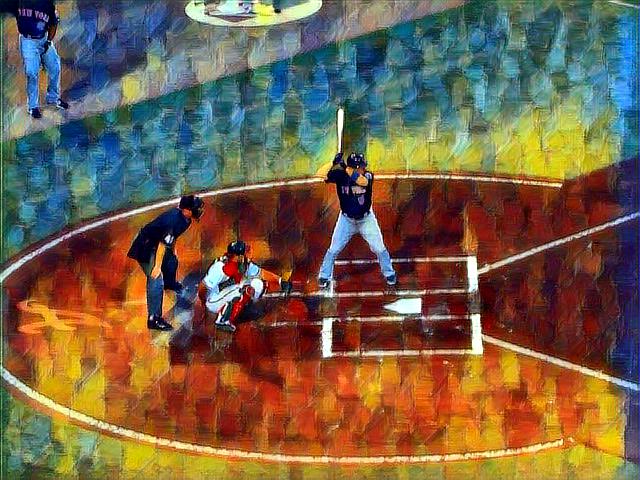} \\
\addstyle{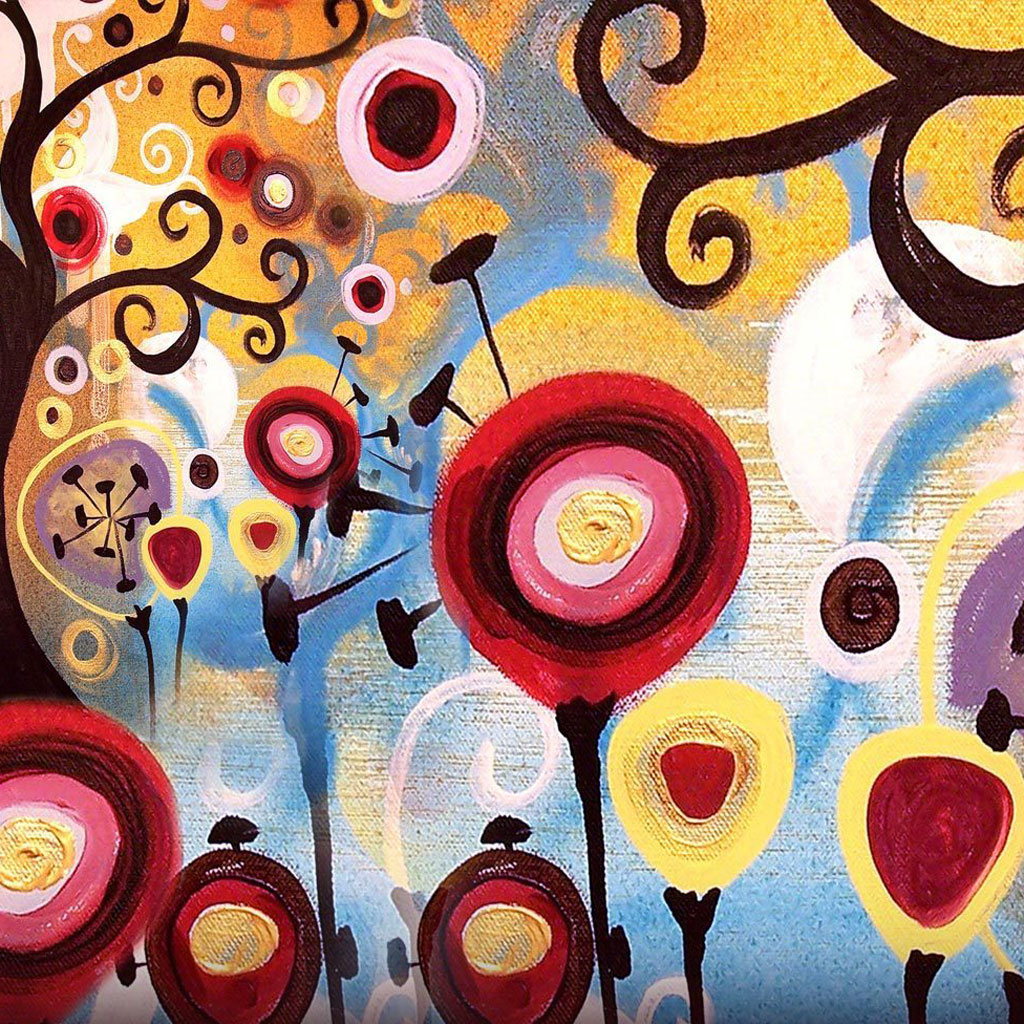} & \addstyle{candy/content} & \addstyle{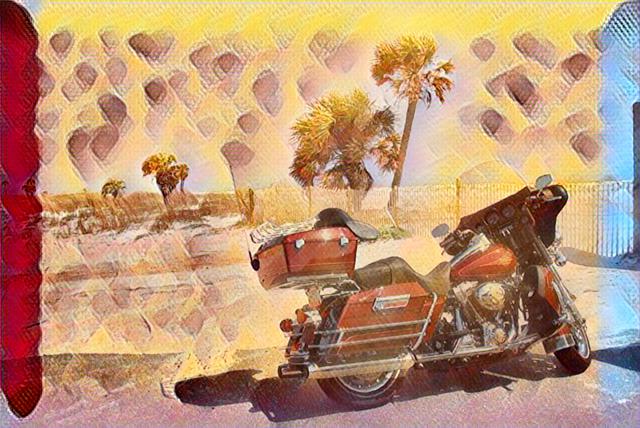}  & \addstyle{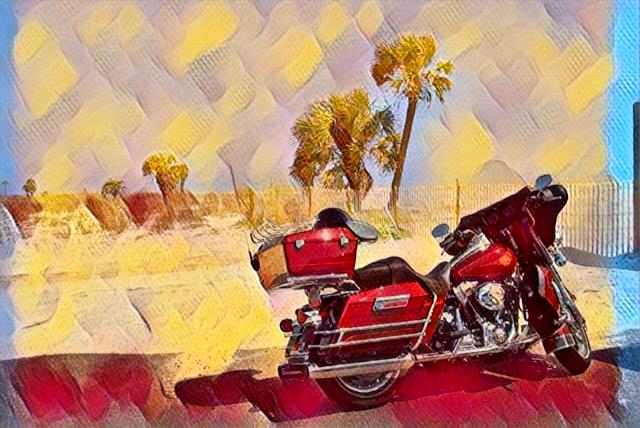} & \addstyle{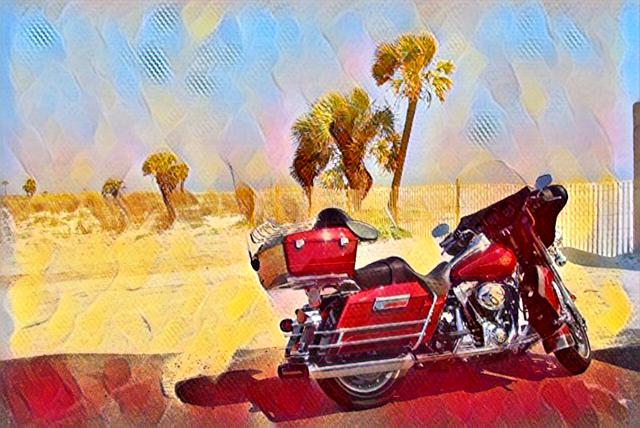} & \addstyle{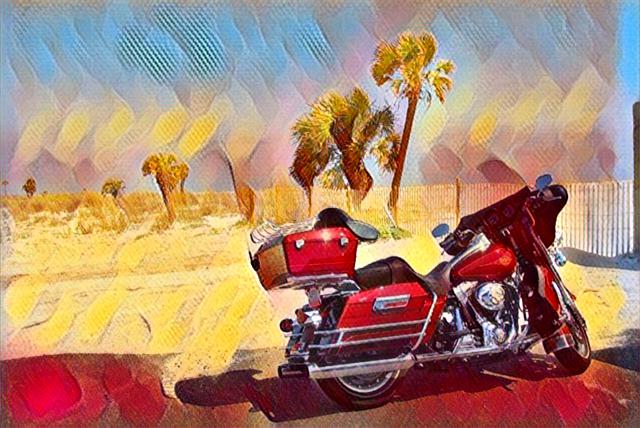} \\
\addstyle{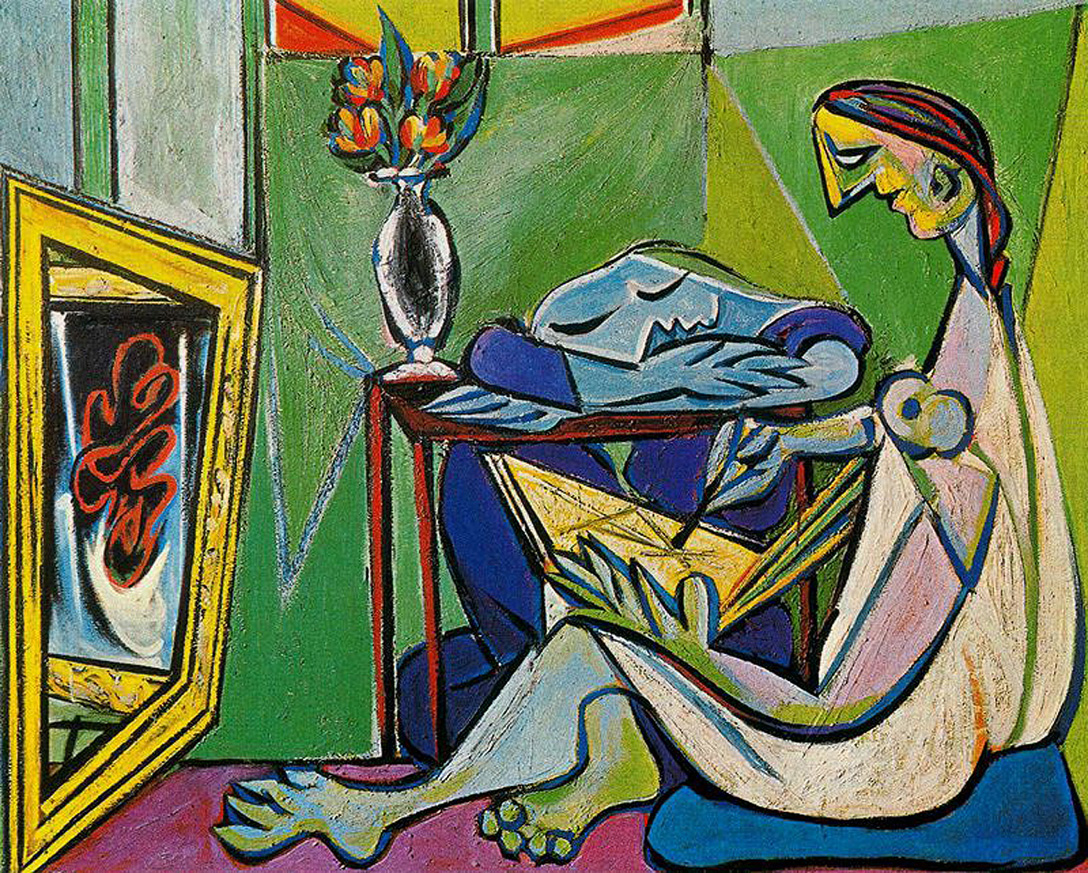} & \addstyle{la-muse/content} & \addstyle{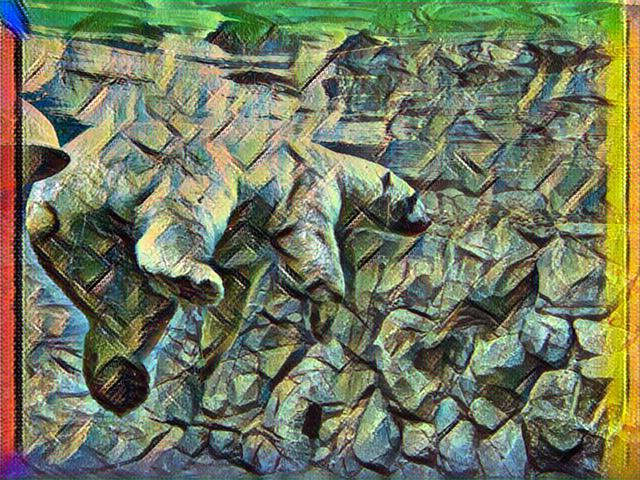}  & \addstyle{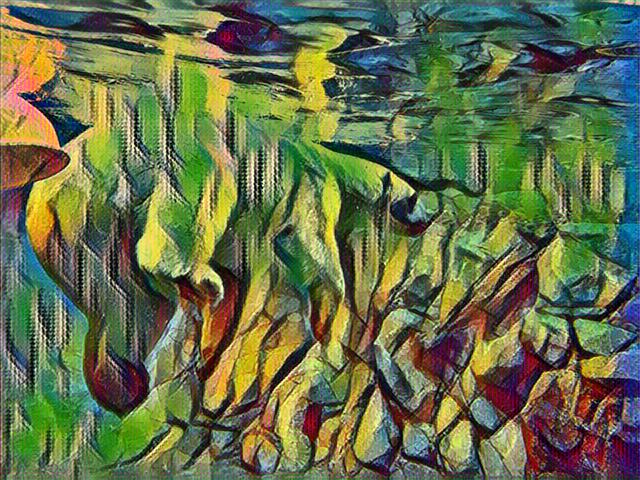} & \addstyle{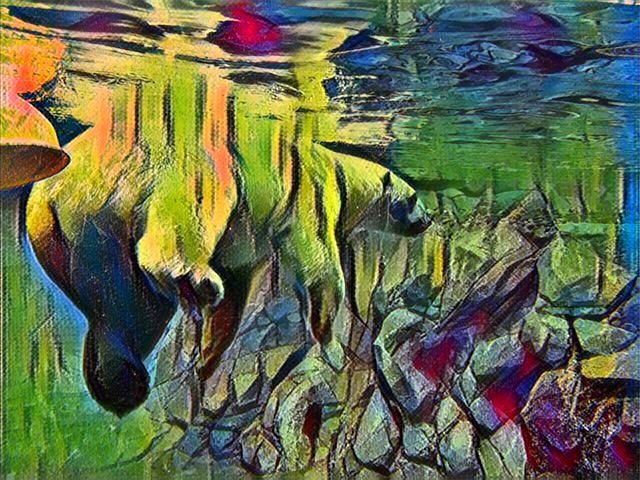} & \addstyle{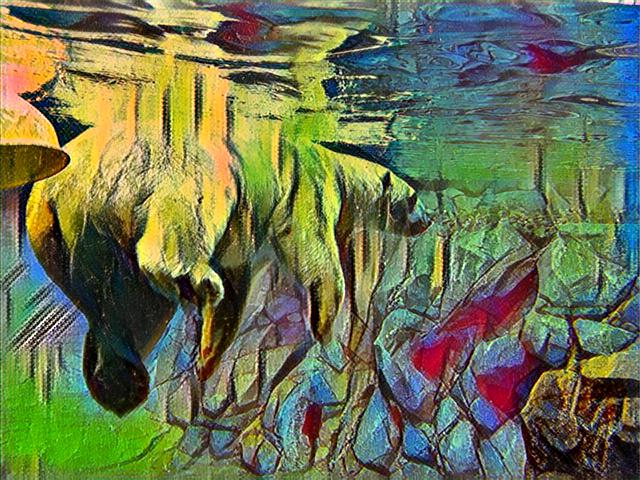} \\
\end{tabular}
\end{center}
\vspace{-1em}
\caption{Example style transfer results. While both BN+SRM and BN+SE improve the stylization quality compared to BN, BN+SRM yields much higher quality which is comparable to IN.
More examples are provided in Figure~\ref{fig:style_transfer_supp}.}
\label{fig:style_transfer}
\end{figure*} 

\paragraph{Stylized-ImageNet.}
We first investigate how SRM handles synthetically increased diversity of styles.
We employ Stylized-ImageNet introduced by \cite{geirhos2019imagenet}, which is constructed by transferring each image in ImageNet to the style of a random painting in the \textit{Painter by Numbers} dataset\footnote{\url{https://www.kaggle.com/c/painter-by-numbers/}} (total 79,434 paintings).
Since the randomly transferred style is irrelevant to the object category, it is a much harder dataset than ImageNet to train on.
We train ResNet-50 based networks on Stylized-ImageNet from scratch\footnote{Although \cite{geirhos2019imagenet} uses ImageNet pretrained networks, we train networks from scratch to focus on the characteristics on Stylized-ImageNet.} following the same training policy as the ImageNet experiment, and report the validation accuracy on Stylized-ImageNet and the original ImageNet in Table~\ref{table:stylized-imagenet}.
SRM not only brings impressive improvements over the baseline and SE on Stylized-ImageNet, but also generalizes better to the original ImageNet.
This supports our claim that SRM learns to suppress the contribution of nuisance styles, which helps the network to concentrate more on meaningful features. 

\paragraph{Multi-Domain Classification.}
We also verify the effectiveness of SRM in tackling natural style variations inherent in different input domains.
We adopt the Office-Home dataset~\cite{venkateswara2017deep} which consists of 15,588 images from 65 categories across 4 heterogeneous domains: Art (Ar), Clip-art (Cl), Product (Pr) and Real-world (Rw).
We combine all training sets of the 4 domains and train domain-agnostic networks based on ResNet-18, following the same setting as the ImageNet experiment except that the networks are trained with a batch size of 64 on 1 GPU.
Table \ref{table:officehome} shows the top-1 accuracy averaged over 5-fold cross validation.
SRM consistently improves the accuracy with significant margins across all domains, which indicates the capability of SRM for alleviating the style discrepancy over different domains.
It also implies the potential of SRM to be utilized in domain adaptation problems~\cite{tzeng2017adversarial,hoffman2018cycada} which entail style disparity between the source and target domains.

\begin{table}
\caption{Top-1 and top-5 accuracy (\%) on the Describable Texture Dataset averaged over 5-fold cross validation. }
\vspace{-1em}
\begin{center}
\begin{tabular}{c|c|c|c|c}
\hline
 & \multicolumn{2}{|c|}{ResNet-32} & \multicolumn{2}{|c}{ResNet-56} \\ \cline{2-5}
 & top-1 & top-5 & top-1 & top-5 \\
\hline
Baseline & 44.96 & 73.85 & 45.46 & 75.54  \\
SE & 45.20 & 75.60 & 48.63 & 77.40  \\
SRM & \textbf{46.50} & \textbf{76.63} & \textbf{50.44} & \textbf{79.37}  \\
\hline
\end{tabular}
\end{center}
\label{table:dtd}
\end{table}

\paragraph{Texture Classification.}
We further evaluate SRM on texture classification using Describable Texture Dataset (DTD)~\cite{cimpoi14describing} which comprises 5,640 images across 47 texture categories such as cracked, bubbly, marbled, etc.
This task offers to assess a different perspective of the network: the ability to extract most textural patterns that elicit visual impressions prior to recognizing objects in images~\cite{cimpoi2016deep}.
We follow the data processing setting of \cite{rebuffi2017learning}, and the same training policy as our CIFAR experiment. 
The results from 5-fold cross validation with ResNet-32 and ResNet-56 baselines are reported in table \ref{table:dtd}, in which SRM achieves outstanding performance improvements. 
It demonstrates that SRM successfully models the importance of individual styles and emphasizes the target textures, enhancing the representational power regarding style attributes.

\begin{figure}
\begin{center}
    \includegraphics[width=\linewidth]{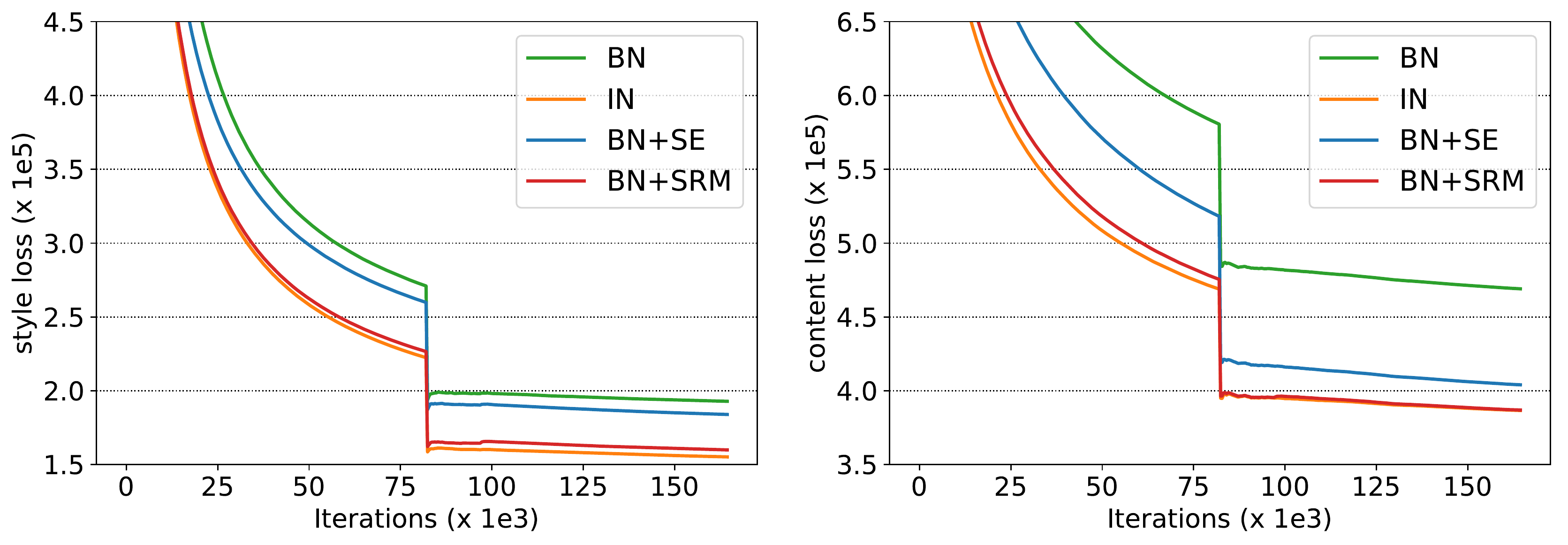}
    \vspace{-1em}
    \caption{Quantitative comparison of style loss (left) and content loss (right) with a style image of \textit{Rain Princess} (the first row in Figure~\ref{fig:style_transfer}).}
    \label{fig:loss}
\end{center}
\end{figure}

\begin{figure*}
\begin{center}
    \includegraphics[width=0.99\textwidth]{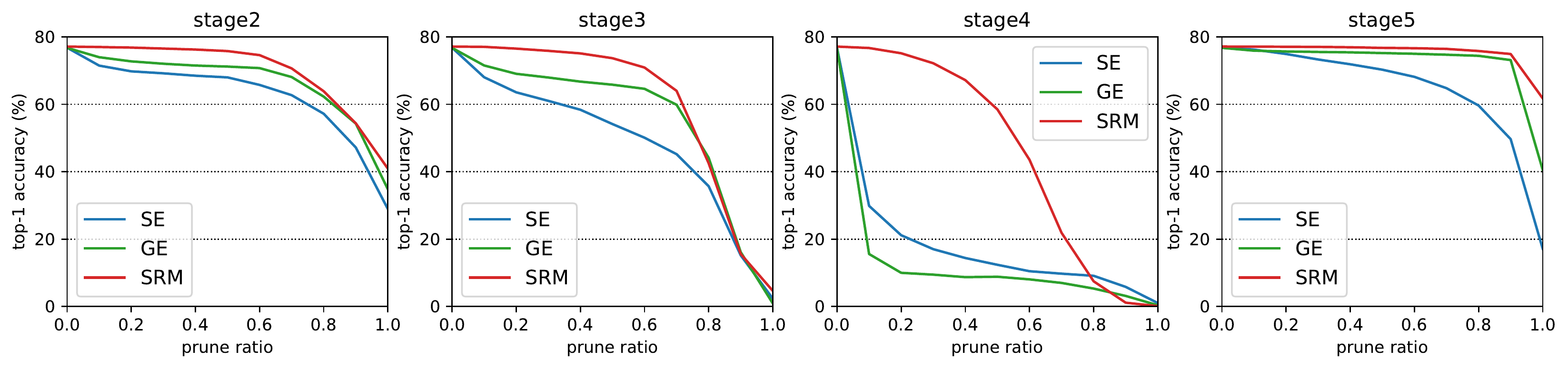}
    \caption{Top-1 validation accuracy of ResNet-50 on ImageNet after pruning channels of each stage according to estimated channel weights.
    Stage 1 is omitted because it consists of a single convolutional layer where a recalibration module is not applied.
    } 
    \label{fig:prune_acc}
\end{center}
\end{figure*}

\subsection{Style Transfer}
We finally examine the benefit of SRM in a generative problem of style transfer.
We utilize a single style feed-forward algorithm~\cite{johnson2016perceptual} implemented in the official PyTorch repository\footnote{\url{https://github.com/pytorch/examples/tree/master/fast\_neural\_style}}.
The networks are trained with content images from the MS-COCO dataset~\cite{lin2014microsoft}, following the default configurations in the original code.

Figure~\ref{fig:loss} depicts the training curves of style and content loss with different recalibration methods.
As reported in the literature~\cite{ulyanov2017improved, nam2018batch}, removing the style from the content image with instance normalization (IN)~\cite{ulyanov2016instance} brings a huge improvement over using the standard batch normalization (BN)~\cite{ioffe2015batch}.
Surprisingly, the BN-based network equipped with SRM (BN+SRM) reaches almost the same level of style/content loss with IN, while the network with SE (BN+SE) exhibits much inferior style/content loss.
This demonstrates the distinct effect of SRM, which mimics the behavior of IN by dynamically suppressing unnecessary styles from input images.
We also show qualitative examples in Figure~\ref{fig:style_transfer}.
Although BN+SE somewhat improves the stylization quality compared to BN, it is still far behind the performance of IN.
In contrast, BN+SRM not only successfully transfers to target style but also better represents the important styles of the content images (e.g. green glass and blue sky), generating competitive results to IN.
Overall, the advantage of SRM is not restricted to discriminative tasks but can be extended to generative frameworks, which remains as future work.
\section{Ablation Study and Analysis} \label{analysis}

In this section, we perform ablation experiments to verify the effectiveness of each component in SRM and in-depth analysis on the behavior of SRM.
As pointed out by Hu \textit{et al.} \cite{hu2018squeeze}, it remains challenging to perform precise theoretical analysis on the feature representation of CNNs.
Instead, we perform an empirical study to gain an insight into the distinguishing role of SRM.

\subsection{Ablation Study} 

\paragraph{Style Pooling.} \label{style_pooling}
We verify the benefit of the proposed style pooling compared to different pooling options.
Throughout the ablation study, we utilize ResNet-50 as a base architecture and address ImageNet classification, following the same procedure as in Section~\ref{image_classification}.
Table \ref{table:pooling_ablation} lists the results of various pooling method fused with style integration operator in our algorithm (except for the baseline).
While each pooling component of SRM (i.e. AvgPool and StdPool) brings meaningful performance improvement, the combination of them further boosts the performance.
We additionally compare our method with MaxPool and the combination of AvgPool and MaxPool proposed in CBAM \cite{woo2018cbam}, which are also outperformed by our style pooling approach.

\paragraph{Style Integration.} \label{style_integration}
We next examine the style integration module which consists of a channel-wise fully connected layer (CFC) followed by a batch normalization layer (BN).
On top of our style pooling operator, we compare CFC with a multi-layer perceptron (MLP) of two fully connected layers (employed in SE) and verify the effect of BN in style integration.
To build MLP on style pooling, we concatenate the style features along the channel axis then apply MLP following the default configuration of SE.
As shown in Table~\ref{table:integration_ablation}, CFC shows better performance than MLP in spite of its simplicity, which highlights the advantage of utilizing channel-wise styles over modeling channel interdependencies.

\begin{table}
\caption{Comparison of different pooling methods on ImageNet validation.}
\vspace{-1em}
\begin{center}
\begin{tabular}{l|c}
\hline
Pooling & top-1 acc.\\
\hline
ResNet-50 (baseline) & 75.89   \\
ResNet-50 + AvgPool & 76.58  \\
ResNet-50 + StdPool & 76.61  \\
ResNet-50 + MaxPool & 75.87  \\
ResNet-50 + AvgPool + MaxPool & 76.35 \\
ResNet-50 + AvgPool + StdPool (SRM) & \textbf{77.13}  \\
\hline
\end{tabular}
\end{center}
\label{table:pooling_ablation}
\end{table}

\begin{table}
\caption{Comparison of different integration methods on ImageNet validation.
\textit{SP}: style pooling,  \textit{MLP}: multi-layer perceptron, \textit{CFC}: channel-wise fully connected layer, \textit{BN}: batch normalization.}
\vspace{-1em}
\begin{center}
\begin{tabular}{l|c}
\hline
Design & top-1 acc.\\
\hline
ResNet-50 + SP + MLP & 76.75  \\
ResNet-50 + SP + MLP + BN & 76.68  \\
ResNet-50 + SP + CFC & 76.91  \\
ResNet-50 + SP + CFC + BN (SRM) & \textbf{77.13}  \\
\hline
\end{tabular}
\end{center}
\label{table:integration_ablation}
\end{table}

\begin{figure}
\begin{center}
\setlength{\tabcolsep}{0.05em}
\begin{tabular}{cc}
\includegraphics[scale=0.25]{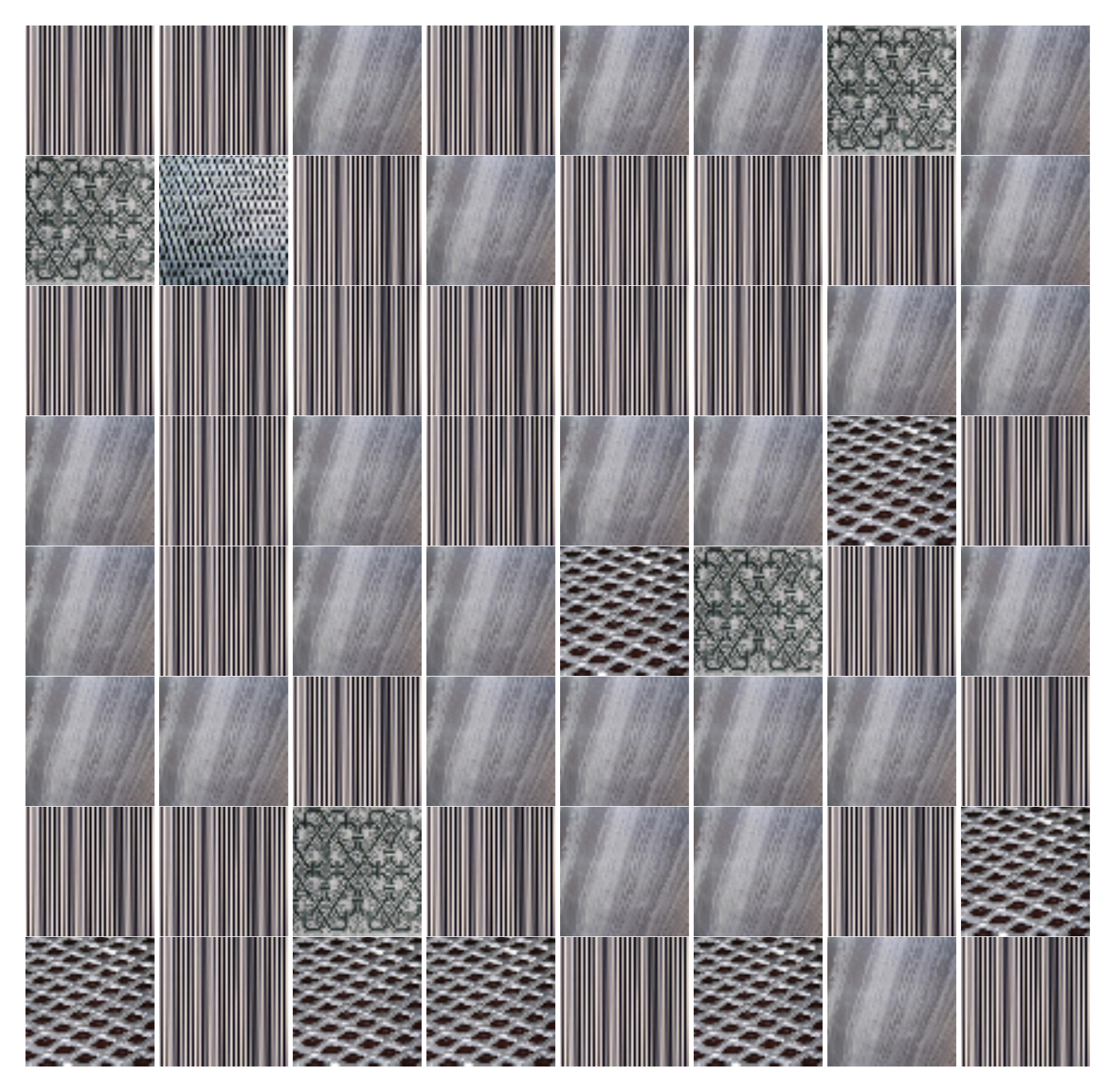} & \includegraphics[scale=0.25]{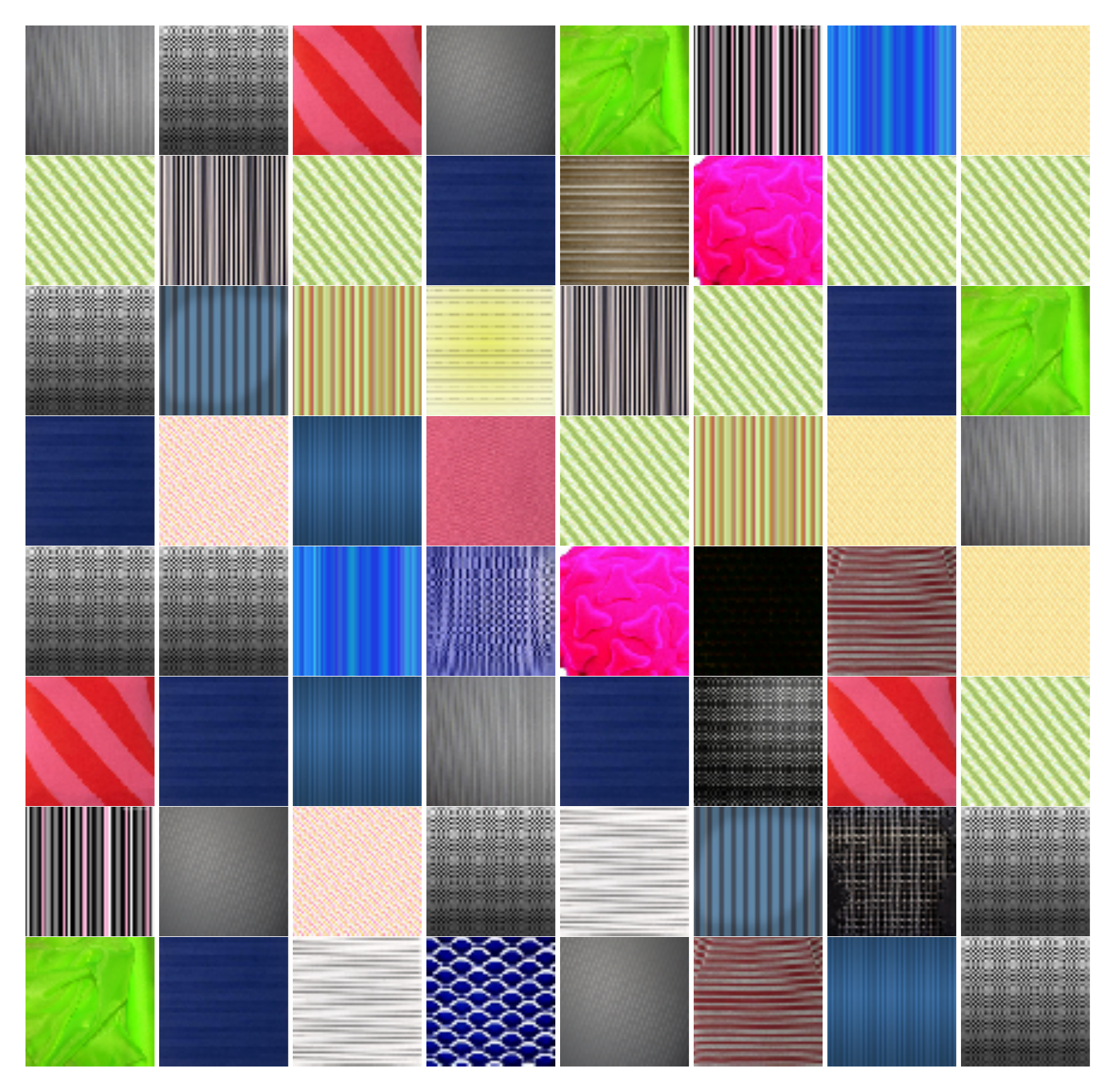} \\
(a) SE & (b) SRM
\end{tabular}
\end{center}
\vspace{-1em}
\caption{The top-activated images for individual channels
in conv2-6 (64 channels) of ResNet-56 on DTD.
More examples are provided in Figure~\ref{fig:dtd_supp}.
}
\label{fig:top_one}
\end{figure}

\begin{figure}
\begin{center}
\setlength{\tabcolsep}{0.05em}
\begin{tabular}{cc}
\includegraphics[height=10.5em]{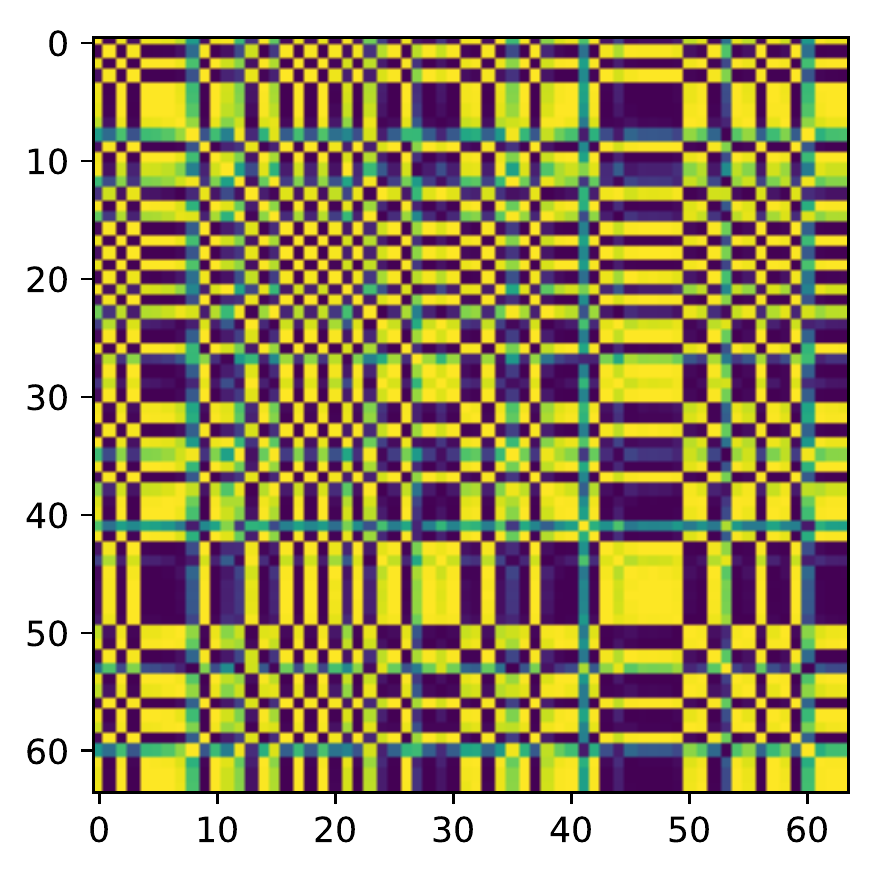} & \includegraphics[height=10.5em]{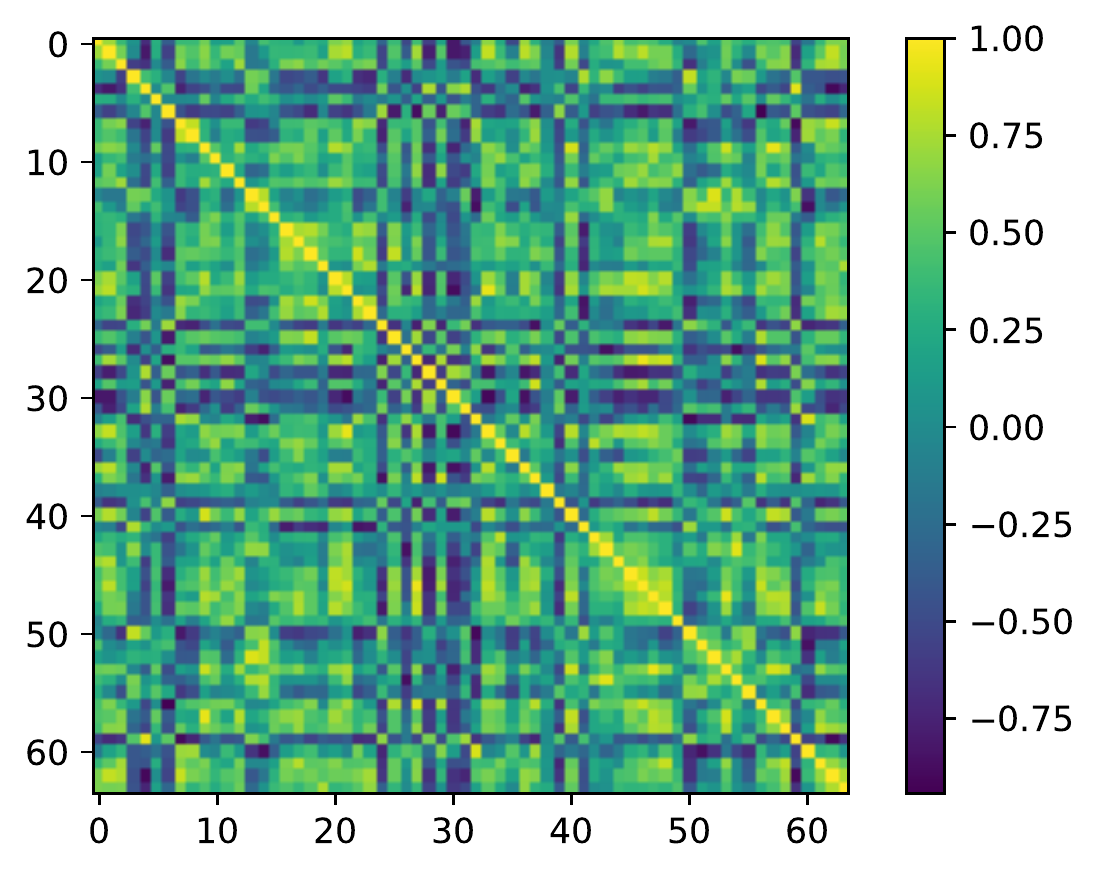} \\
(a) SE & (b) SRM
\end{tabular}
\end{center}
\vspace{-1em}
\caption{Visualization of the correlation matrix between the channel weights in conv2-6 (64$\times$64) of ResNet-56 on DTD.
More examples are provided in Figure~\ref{fig:dtd_supp}.
}
\label{fig:corr_mat}
\end{figure}

\subsection{Channel Pruning}
SRM learns to adaptively predict the channel-wise importance of feature maps.
In this regard, we evaluate the validity of the feature importance learned by SRM through channel pruning of ResNet-50 on ImageNet classification.
Given an input image in the validation set, we sort the channel weights of each residual block at certain stage in ascending order. Then, we select the channels to be pruned in order according to a prune ratio.
Since each pruned channel is filled with zero, the amount of information to be passed decreases as the prune ratio increases.
In an extreme case where the prune ratio is equal to one, the input feature maps directly pass through an identity mapping ignoring the residual block.

We compare the validation accuracy when channel pruning is applied to SE, GE, and SRM at different stages and report the results in Figure~\ref{fig:prune_acc}.
The accuracy is mostly preserved during the early phase of the pruning process but it quickly drops after a certain prune ratio.
Throughout all stages, the accuracy drops noticeably slower in SRM compared to SE and GE, which implies that SRM learns better relative importance of channels than other methods.
Note that SRM predicts channel importance solely based on style context, which may provide an insight into how the network utilizes the style of an image in its decision making process.

\subsection{Difference between SRM and SE Block}
Although the proposed SRM shares similar aspects of feature recalibration with the SE block, we observe the characteristics of SRM is far distinct from SE throughout the experiments.
To further understand their representational difference, we visualize the features learned by each method through seeking the images that leads to the highest channel weights.
We record the channel weights for each validation image obtained by SE-ResNet-56 and SRM-ResNet-56 trained on DTD.
Figure~\ref{fig:top_one} shows the top-activated images for individual channels in conv2-6 among the entire validation set. While SE results in highly overlapped images across channels, SRM yields a greater diversity of top-activated images.
This implies SRM allows lower correlation between channel weights compared to the SE block, which leads us to the following exploration.

Figure~\ref{fig:corr_mat} depicts the correlation matrix between channel weights produced by SE and SRM.
As expected, there exists high correlation between the channel weights in the SE block, but SRM exhibits lower correlation between channels (in terms of the total sum of squared correlation coefficients throughout the whole network, SRM shows almost three times smaller numerical value of 143,909 than SE's 420,509). 
In addition, the conspicuous grid pattern in SE's correlation matrix implies that groups of channels are turned on or off synchronously, whereas SRM tends to encourage decorrelation between channels.
Our comparison between SE and SRM suggests that they target quite different perspectives of feature representations to enhance performance, which is worth future investigation.


\section{Conclusion}
In this work, we present Style-based Recalibration Module (SRM), a lightweight architectural unit that dynamically recalibrates feature responses based on style importance.
By incorporating the styles into feature maps, it effectively enhances the representational power of a CNN.
Our experiments on general object classification demonstrate that simply inserting SRM into standard CNN architectures such as ResNet boosts the performance of network.
Furthermore, we verify the significance of SRM in controlling the contribution of styles through various style-related tasks.
While most previous works utilized styles in image generation frameworks, SRM is designed to harness the latent ability of style information in more general vision tasks.
We hope our work sheds light on better exploiting styles into designing a CNN architecture in a wide range of applications. 

\begin{figure*}
\newcommand{\addstyle}[1]{\includegraphics[width=6em, height=6em]{supp_figure/style_transfer/#1.jpg}}
\newcommand{\addfile}[1]{\includegraphics[width=6em, height=6em]{figure/style_transfer/#1.jpg}}

\begin{center}
\setlength{\tabcolsep}{0.1em}
\begin{tabular}{cccccc}
Style & Content & BN  & BN+SE & BN+SRM & IN \\
\addfile{rain-princess/style} & \addstyle{rain-princess/case1/content} & \addstyle{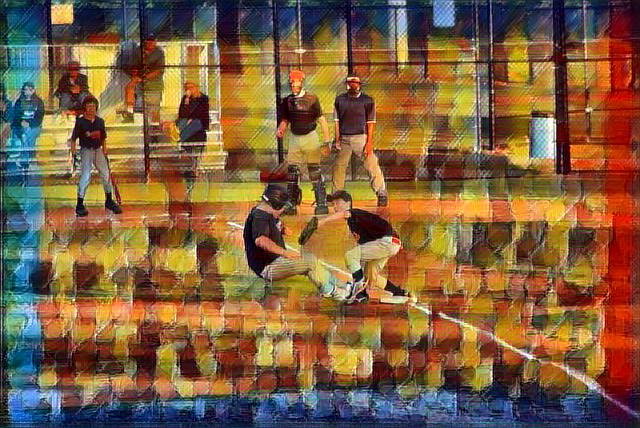}  & \addstyle{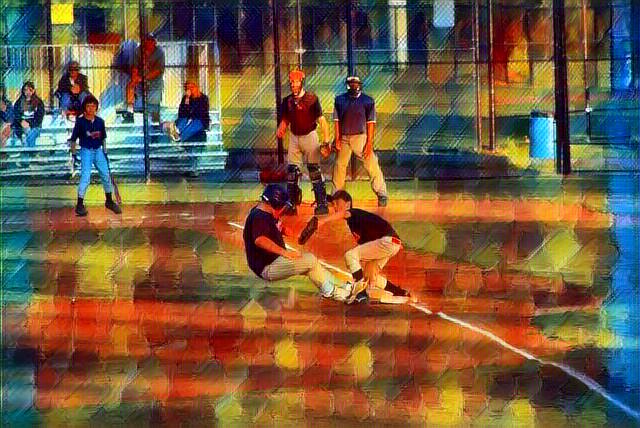} & \addstyle{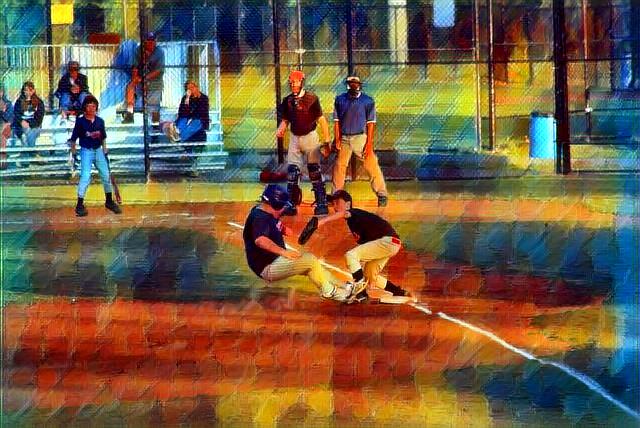} & \addstyle{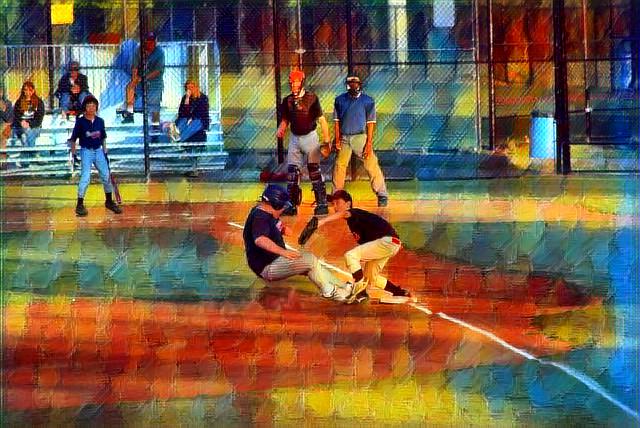} \\
 & \addstyle{rain-princess/case2/content} & \addstyle{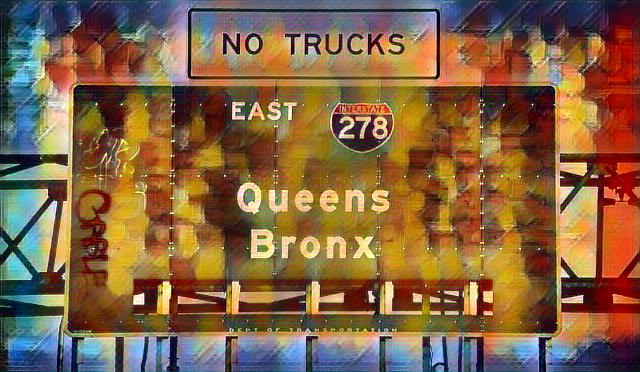}  & \addstyle{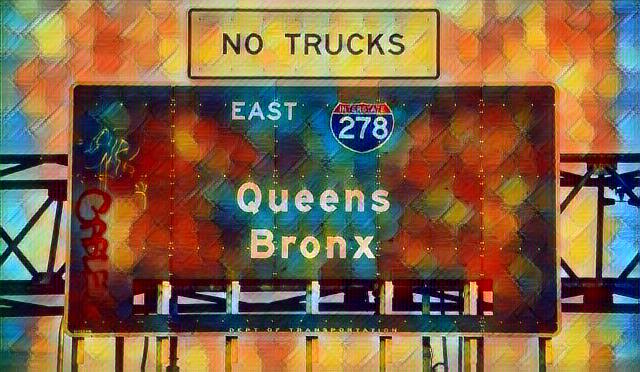} & \addstyle{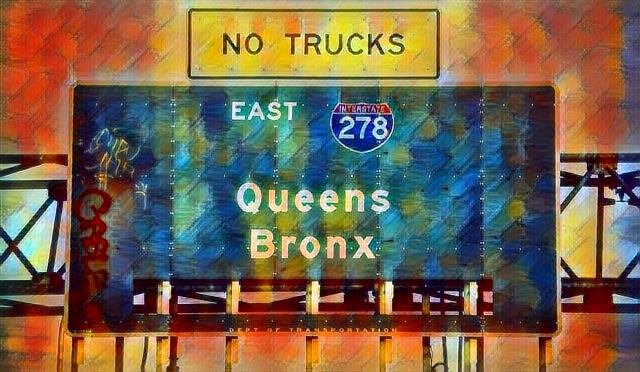} & \addstyle{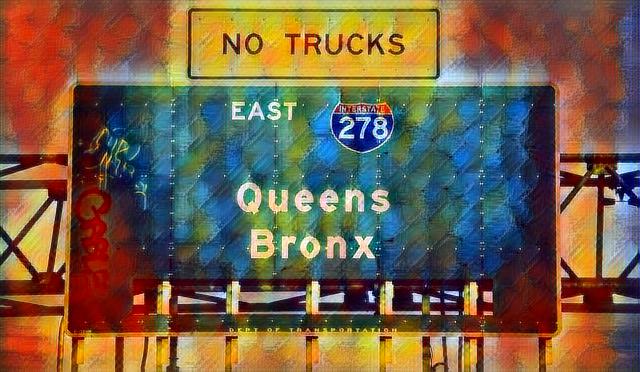} \\
 & \addstyle{rain-princess/case3/content} & \addstyle{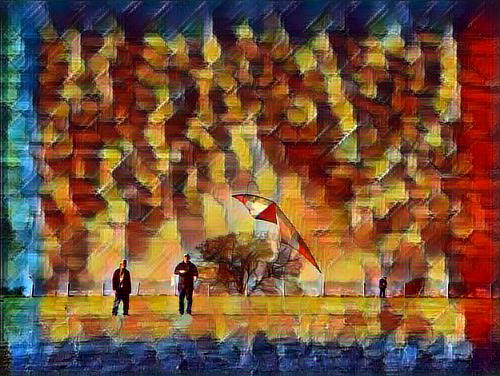}  & \addstyle{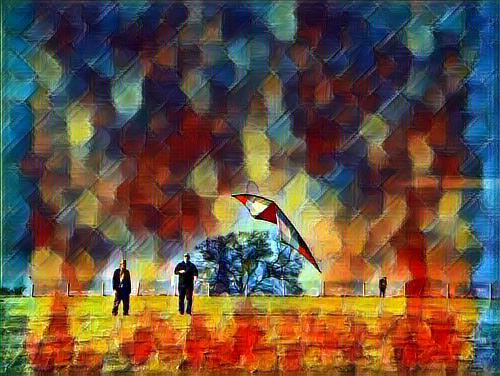} & \addstyle{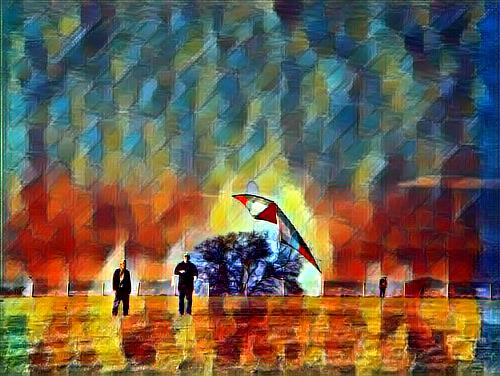} & \addstyle{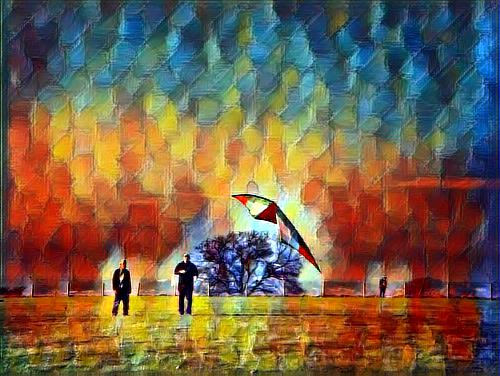} \\
\addfile{candy/style} & \addstyle{candy/case1/content} & \addstyle{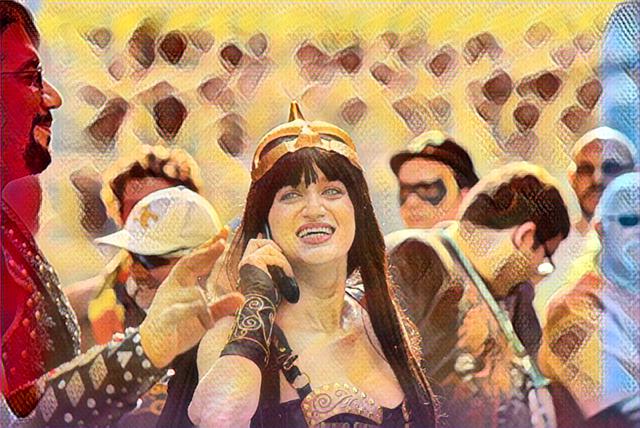}  & \addstyle{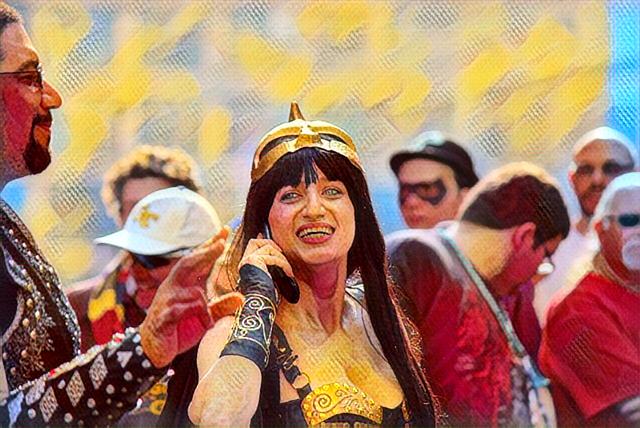} & \addstyle{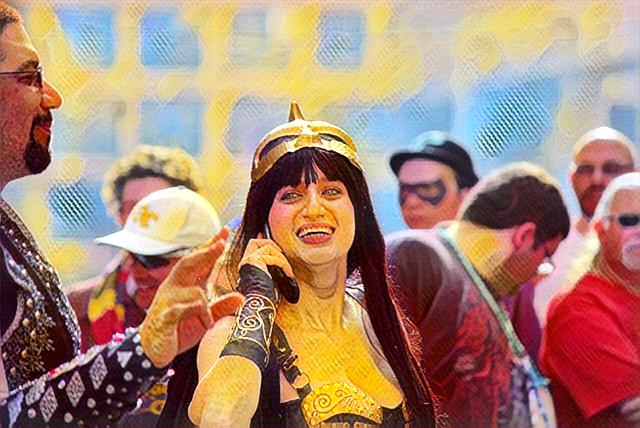} & \addstyle{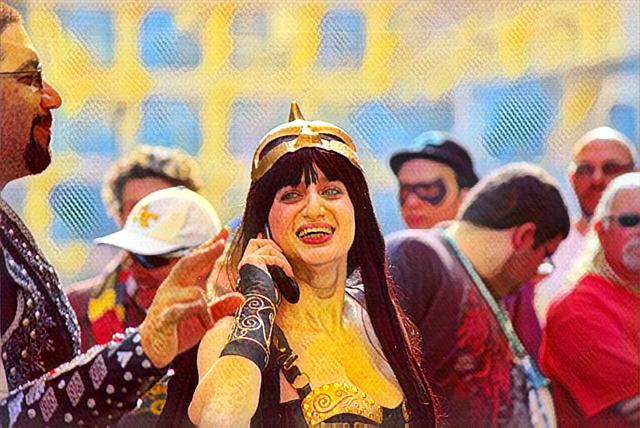} \\
 & \addstyle{candy/case2/content} & \addstyle{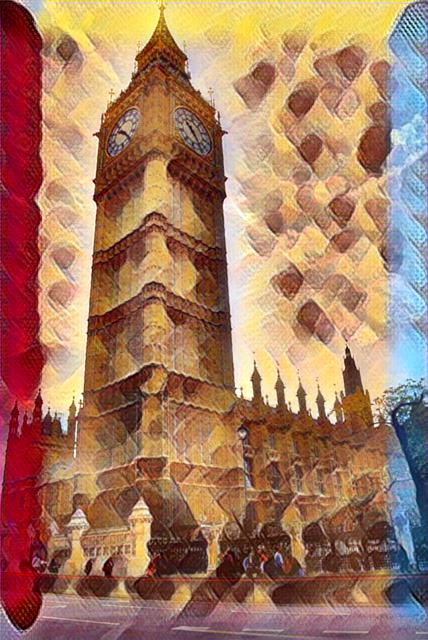}  & \addstyle{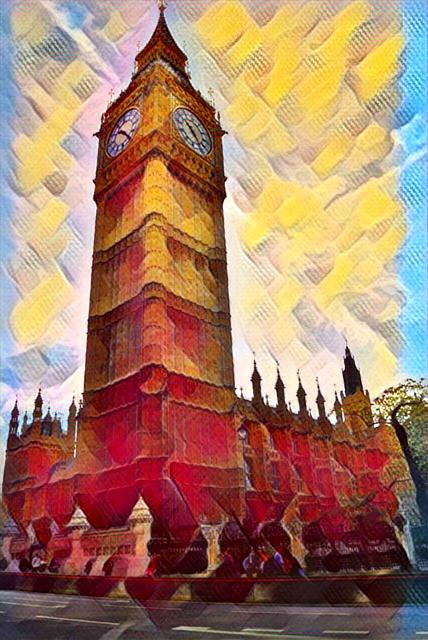} & \addstyle{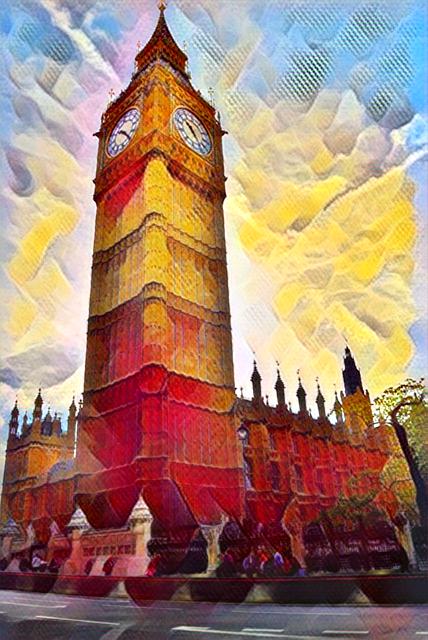} & \addstyle{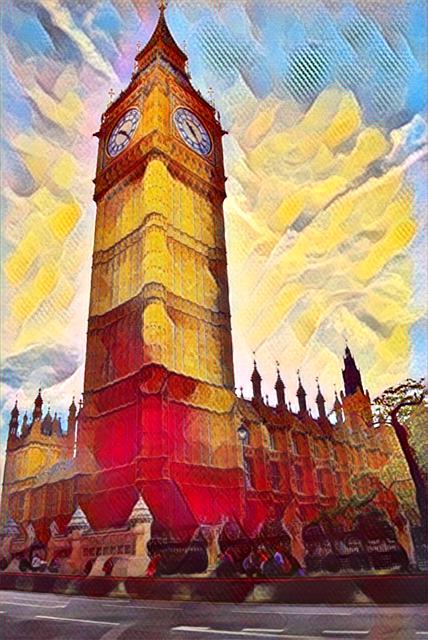} \\
 & \addstyle{candy/case3/content} & \addstyle{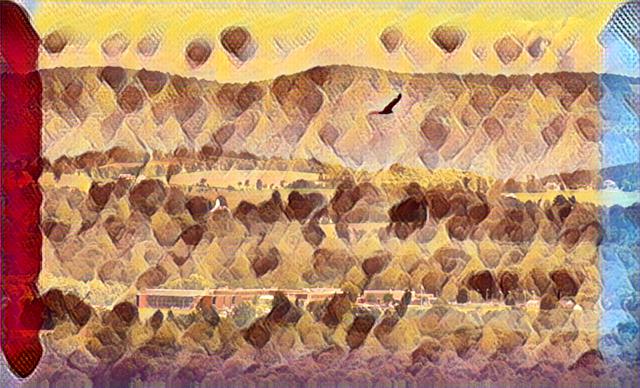}  & \addstyle{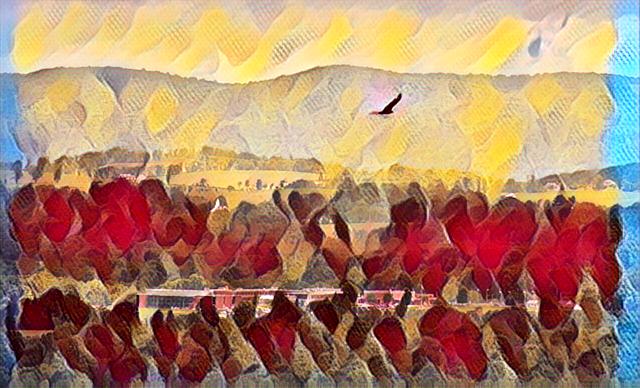} & \addstyle{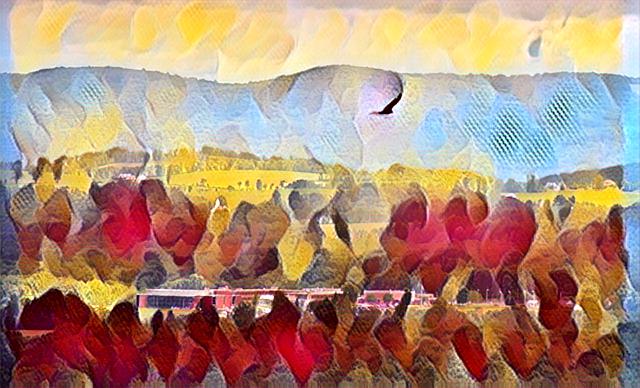} & \addstyle{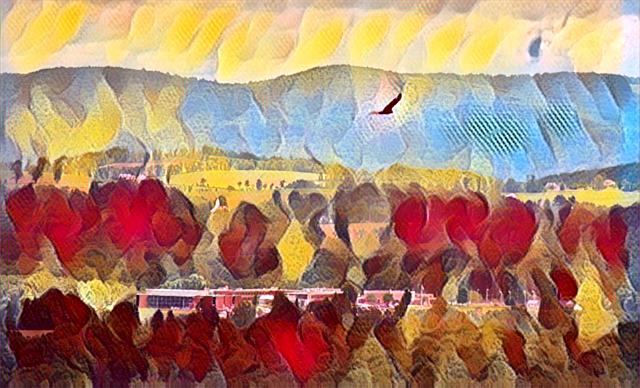} \\

\addfile{la-muse/style} & \addstyle{la-muse/case1/content} & \addstyle{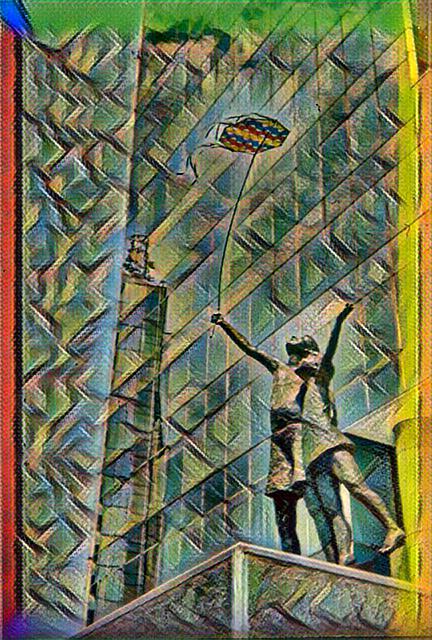}  & \addstyle{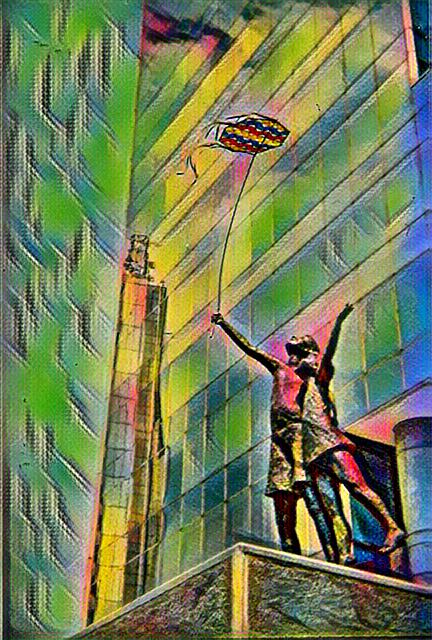} & \addstyle{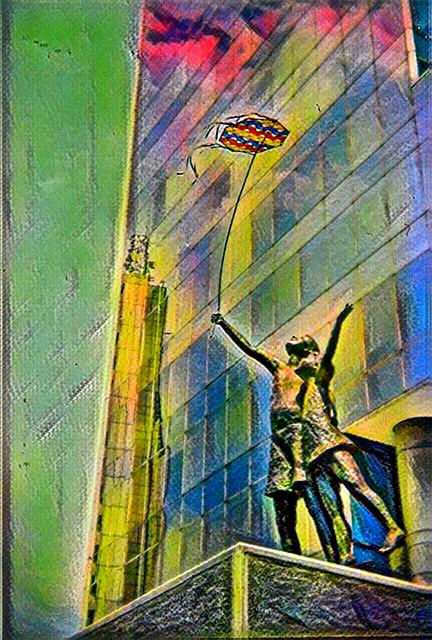} & \addstyle{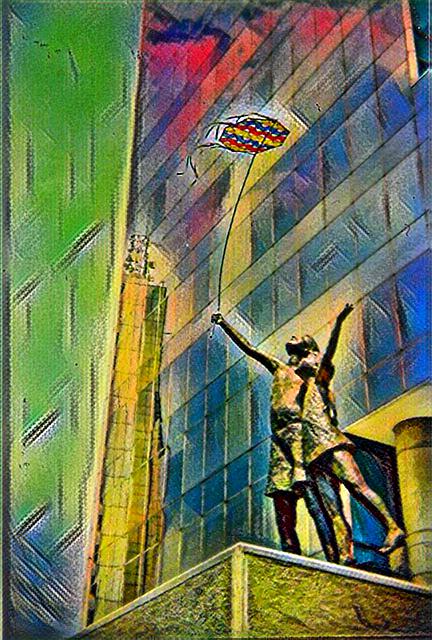} \\
 & \addstyle{la-muse/case2/content} & \addstyle{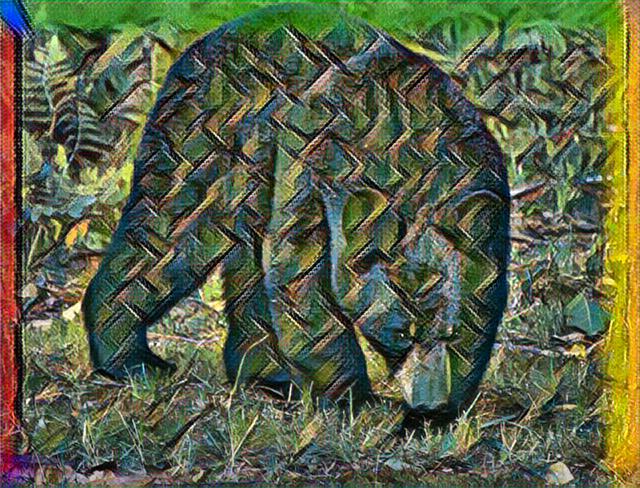}  & \addstyle{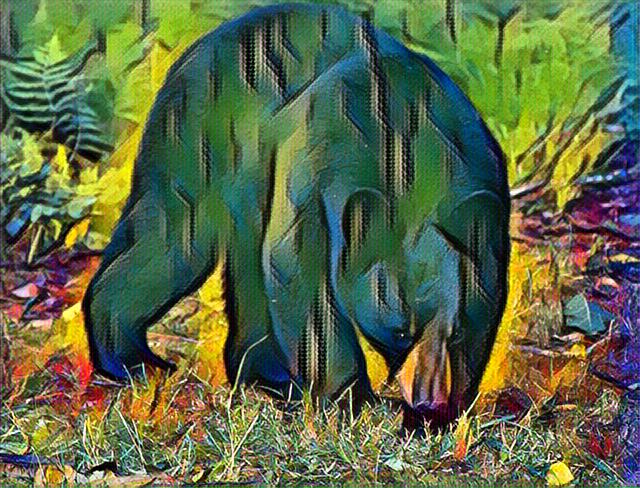} & \addstyle{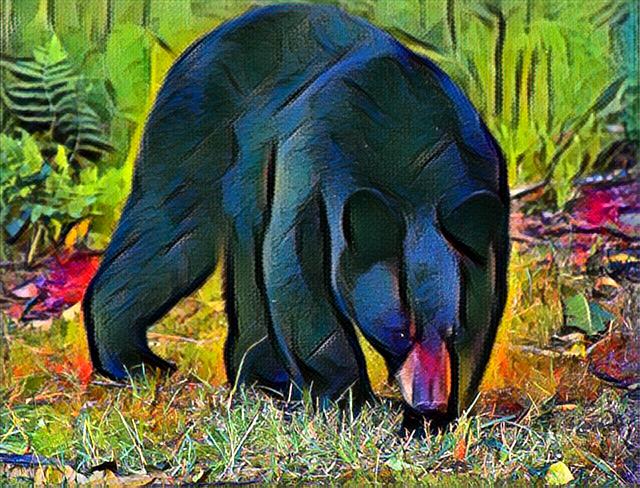} & \addstyle{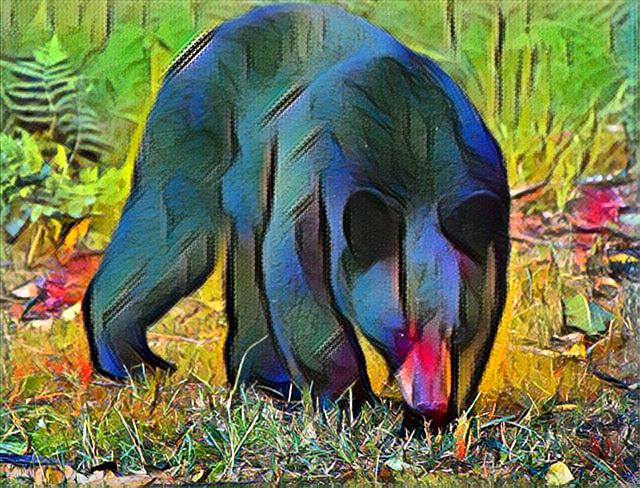} \\
  & \addstyle{la-muse/case3/content} & \addstyle{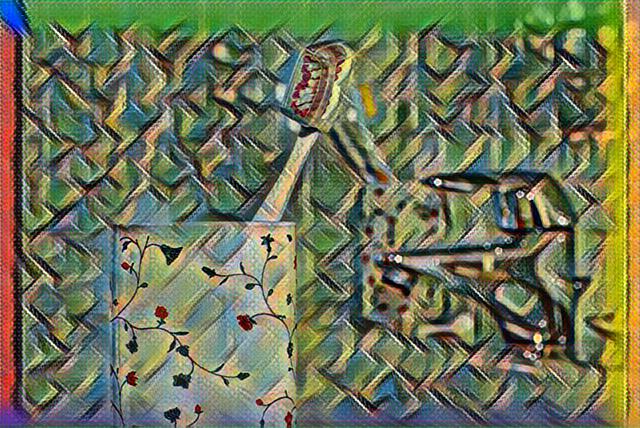}  & \addstyle{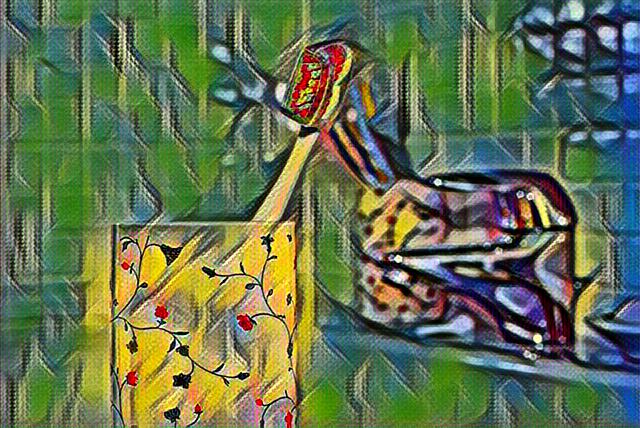} & \addstyle{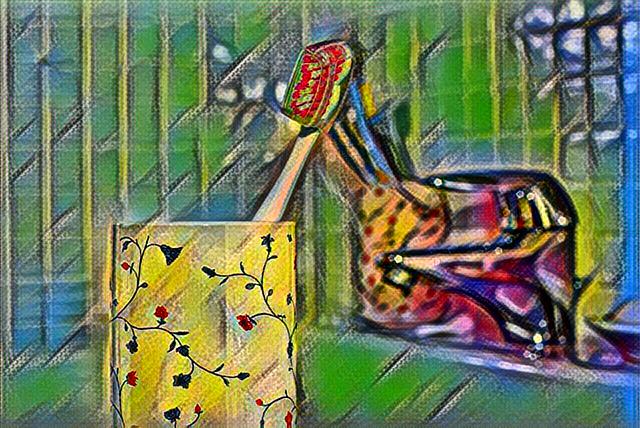} & \addstyle{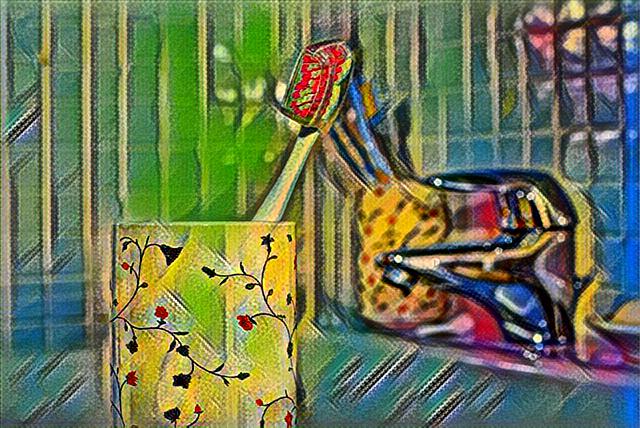} \\

\end{tabular}
\end{center}
\caption{Additional examples of style transfer.
While BN results in vague boundaries between areas along with severe artifacts and BN+SE alleviates them to some degree, BN+SRM yields considerably higher stylization quality which is comparable to IN.
}
\label{fig:style_transfer_supp}
\end{figure*}

\begin{figure*}
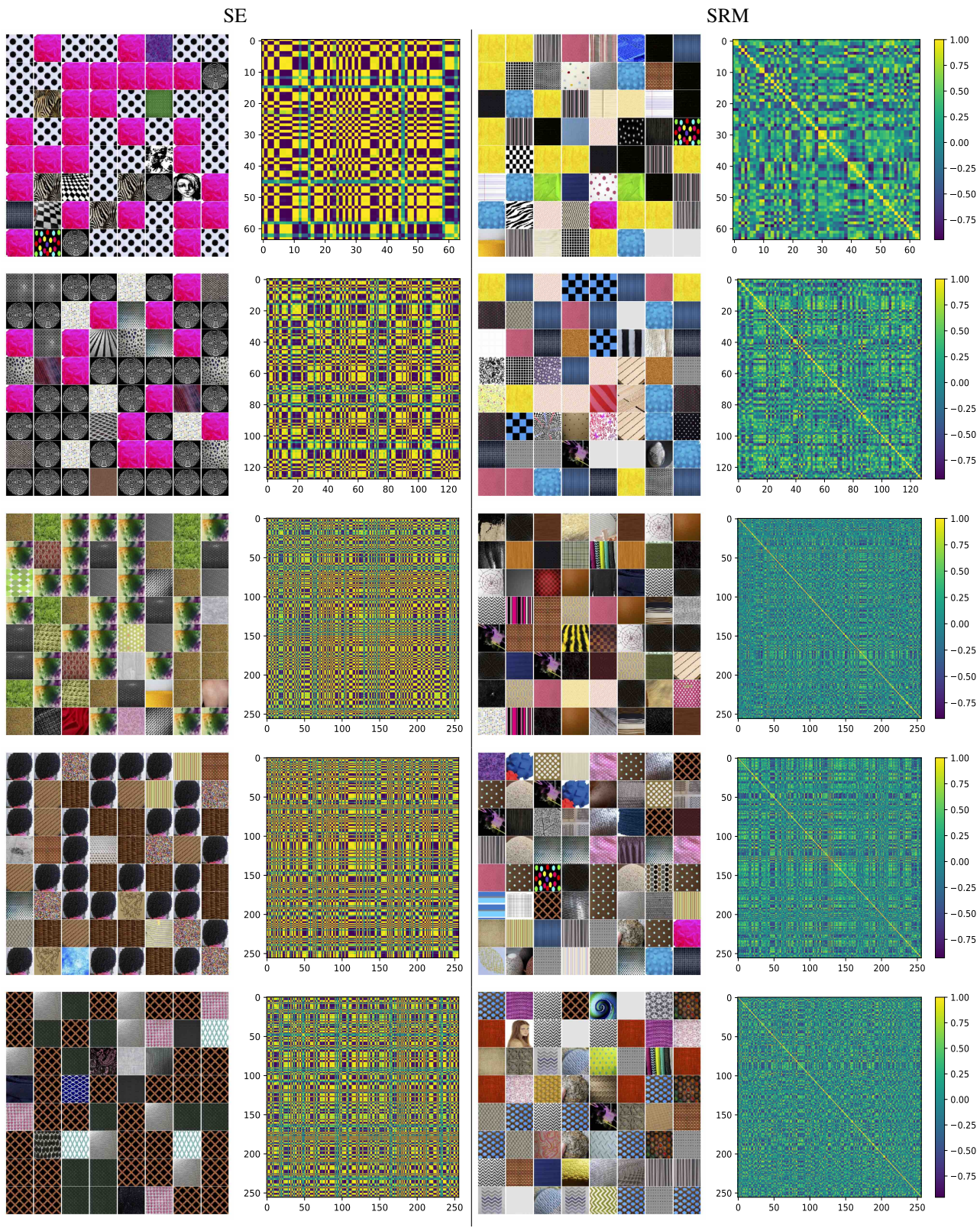

\newcommand{\addtop}[1]{\includegraphics[width=11.5em, height=11.5em]{{supp_figure/dtd/top-1/layer#1_top_one}.jpg}}
\newcommand{\addcorrse}[1]{\includegraphics[width=11.5em, height=11.5em]{{supp_figure/dtd/corr/layer#1_se_corr}.pdf}}
\newcommand{\addcorrsrm}[1]{\includegraphics[width=13.5em, height=11.5em]{{supp_figure/dtd/corr/layer#1_sam_corr}.pdf}}

\begin{center}
\setlength{\tabcolsep}{0.1em}
\begin{tabular}{cc|cc}
   \multicolumn{2}{c}{SE} & \multicolumn{2}{c}{SRM}  \\
 \addtop{1.5_se}  & \addcorrse{1.5} & \addtop{1.5_sam} & \addcorrsrm{1.5} \\
 \addtop{2.6_se}  & \addcorrse{2.6} & \addtop{2.6_sam} & \addcorrsrm{2.6} \\
 \addtop{3.4_se}  & \addcorrse{3.4} & \addtop{3.4_sam} & \addcorrsrm{3.4} \\
 \addtop{3.5_se}  & \addcorrse{3.5} & \addtop{3.5_sam} & \addcorrsrm{3.5} \\
 \addtop{3.6_se}  & \addcorrse{3.6} & \addtop{3.6_sam} & \addcorrsrm{3.6} \\

\end{tabular}
\end{center}
\caption{The top-activated images of the first 64 channels in channel weights and the correlation matrix between channel weights of ResNet-56 on Describable Texture Dataset.
Each row (from top to bottom) corresponds to conv2\_5, conv3\_6, conv4\_4, conv4\_5, and conv4\_6, respectively.}
\label{fig:dtd_supp}
\end{figure*}


{\small
\bibliographystyle{ieee}

}

\end{document}